\documentclass[10pt,twocolumn,letterpaper]{article}

\usepackage{cvpr}
\usepackage{times}
\usepackage{epsfig}
\usepackage{graphicx}
\usepackage{amsmath}
\usepackage{amssymb}

\usepackage{times}
\usepackage{epsfig}
\usepackage{graphicx}
\usepackage{amsmath}
\usepackage{amssymb}

\usepackage{caption}

\usepackage{makecell}
\usepackage{booktabs}
\usepackage{multirow}
\usepackage{siunitx}

\usepackage[T1]{fontenc}
\usepackage{fix-cm}

\newcolumntype{?}{!{\vrule width 1pt}}
\newcolumntype{C}[1]{>{\centering}m{#1}}

\newcolumntype{X}{@{\hskip\tabcolsep\vrule width 1.5pt\hskip\tabcolsep}}

\newcommand{\HfL}{\textsc{HfL}\xspace}

\newcommand{\myfigurefivecol}[1]{
\begin{minipage}[b]{.18\textwidth}
\includegraphics[width=1.03\linewidth]{#1}
\end{minipage}
}

\newcommand{\myfigurefivecolcaption}[2]{
\begin{minipage}[b]{.18\textwidth}
\includegraphics[width=1.03\linewidth]{#1}
\caption{{\small {#2}}}
\end{minipage}
}

\newcommand{\myfigurethreecol}[1]{
\begin{minipage}[b]{.14\textwidth}
\includegraphics[width=1.1\linewidth]{#1}
\end{minipage}
}

\usepackage[pagebackref=true,breaklinks=true,letterpaper=true,colorlinks,bookmarks=false]{hyperref}

 \cvprfinalcopy % *** Uncomment this line for the final submission

\def\cvprPaperID{1172} % *** Enter the CVPR Paper ID here

\ifcvprfinal\fi
\begin{document}

%%%%%%%%% TITLE
\title{Semantic Segmentation with Boundary Neural Fields}

\author{Gedas Bertasius\\
University of Pennsylvania\\
{\tt\small gberta@seas.upenn.edu}
% For a paper whose authors are all at the same institution,
% omit the following lines up until the closing ``}''.
% Additional authors and addresses can be added with ``\and'',
% just like the second author.
% To save space, use either the email address or home page, not both
\and
Jianbo Shi\\
University of Pennsylvania\\
{\tt\small jshi@seas.upenn.edu}
\and
Lorenzo Torresani\\
Dartmouth College\\
{\tt\small lt@dartmouth.edu}
}

\maketitle
%\thispagestyle{empty}

%%%%%%%%% ABSTRACT
\begin{abstract}

The state-of-the-art in semantic segmentation is currently represented by fully convolutional networks (FCNs). However, FCNs use large receptive fields and many pooling layers, both of which cause blurring and low spatial resolution in the deep layers. As a result FCNs tend to produce segmentations that are poorly localized around object boundaries. Prior work has attempted to address this issue in post-processing steps, for example using a color-based CRF on top of the FCN predictions. However, these approaches require additional parameters and low-level features that are difficult to tune and integrate into the original network architecture. Additionally, most CRFs use color-based pixel affinities, which are not well suited for semantic segmentation and lead to spatially disjoint predictions.

To overcome these problems, we introduce a Boundary Neural Field (BNF), which is a global energy model integrating FCN predictions with boundary cues. The boundary information is used to enhance semantic segment coherence and to improve object localization. Specifically, we first show that the convolutional filters of semantic FCNs provide good features for boundary detection. We then employ the predicted boundaries to define pairwise potentials in our energy. Finally, we show that our energy decomposes semantic segmentation into multiple binary problems, which can be relaxed for efficient global optimization. We report extensive experiments demonstrating that minimization of our global boundary-based energy yields results superior to prior globalization methods, both quantitatively  as well as qualitatively.

\end{abstract}

%\captionsetup{labelformat=empty}

\begin{figure}
\centering

%\myfigurethreecol{./paper_figures/intro/2008_000003.pdf}
%\myfigurethreecol{./paper_figures/intro/2008_000003_fc8.pdf}
%\myfigurethreecol{./paper_figures/intro/2008_000003_crf.pdf}

\myfigurethreecol{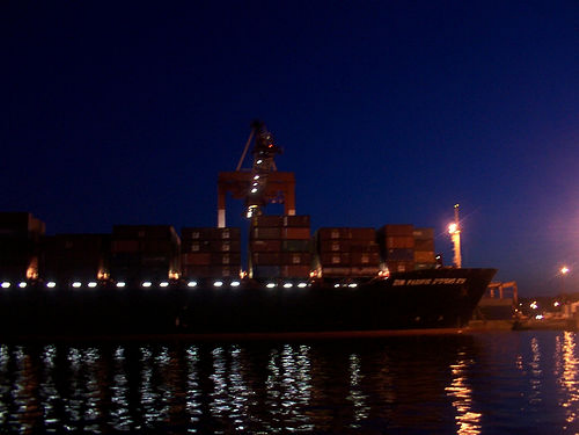}
\myfigurethreecol{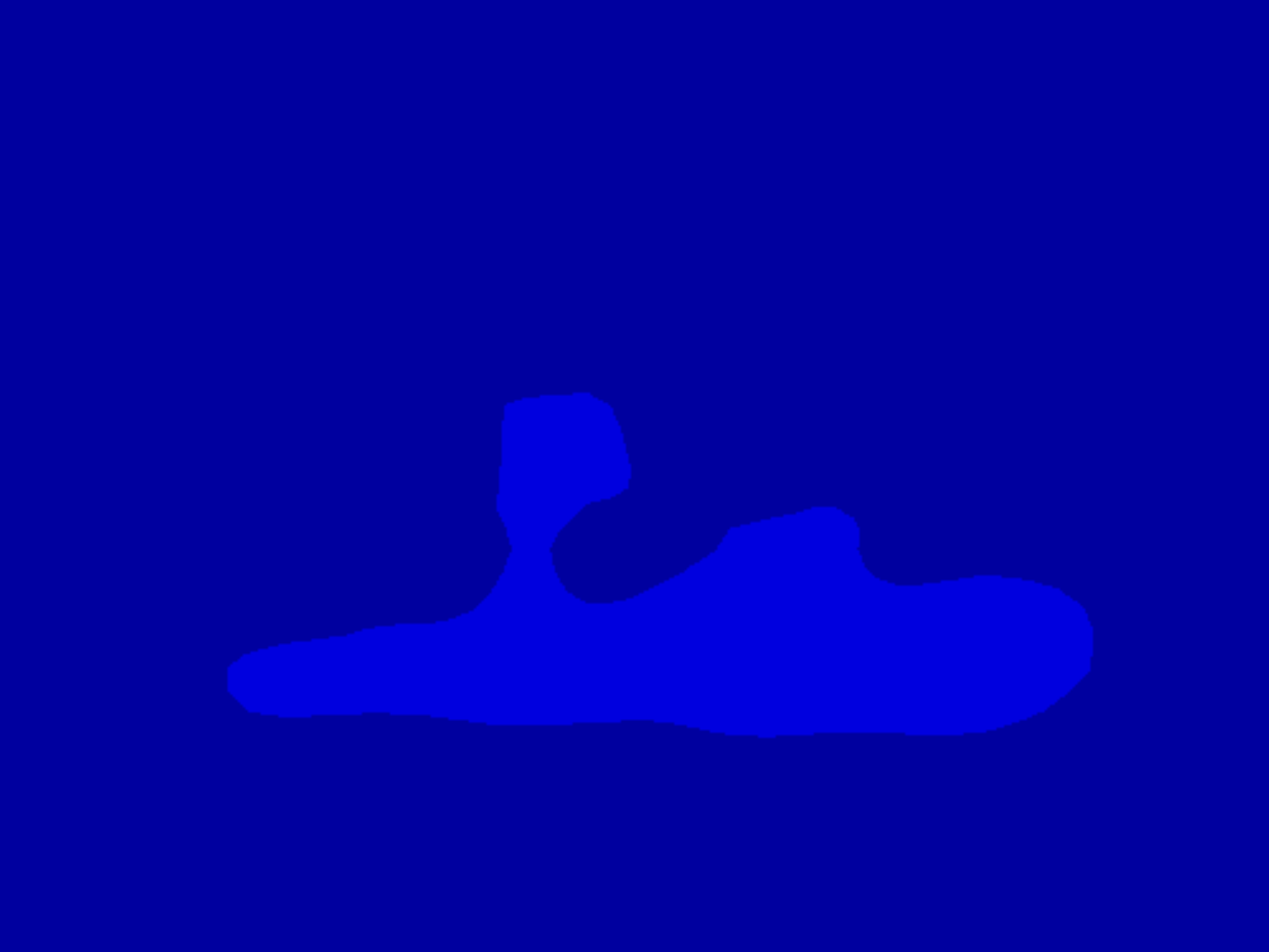}
\myfigurethreecol{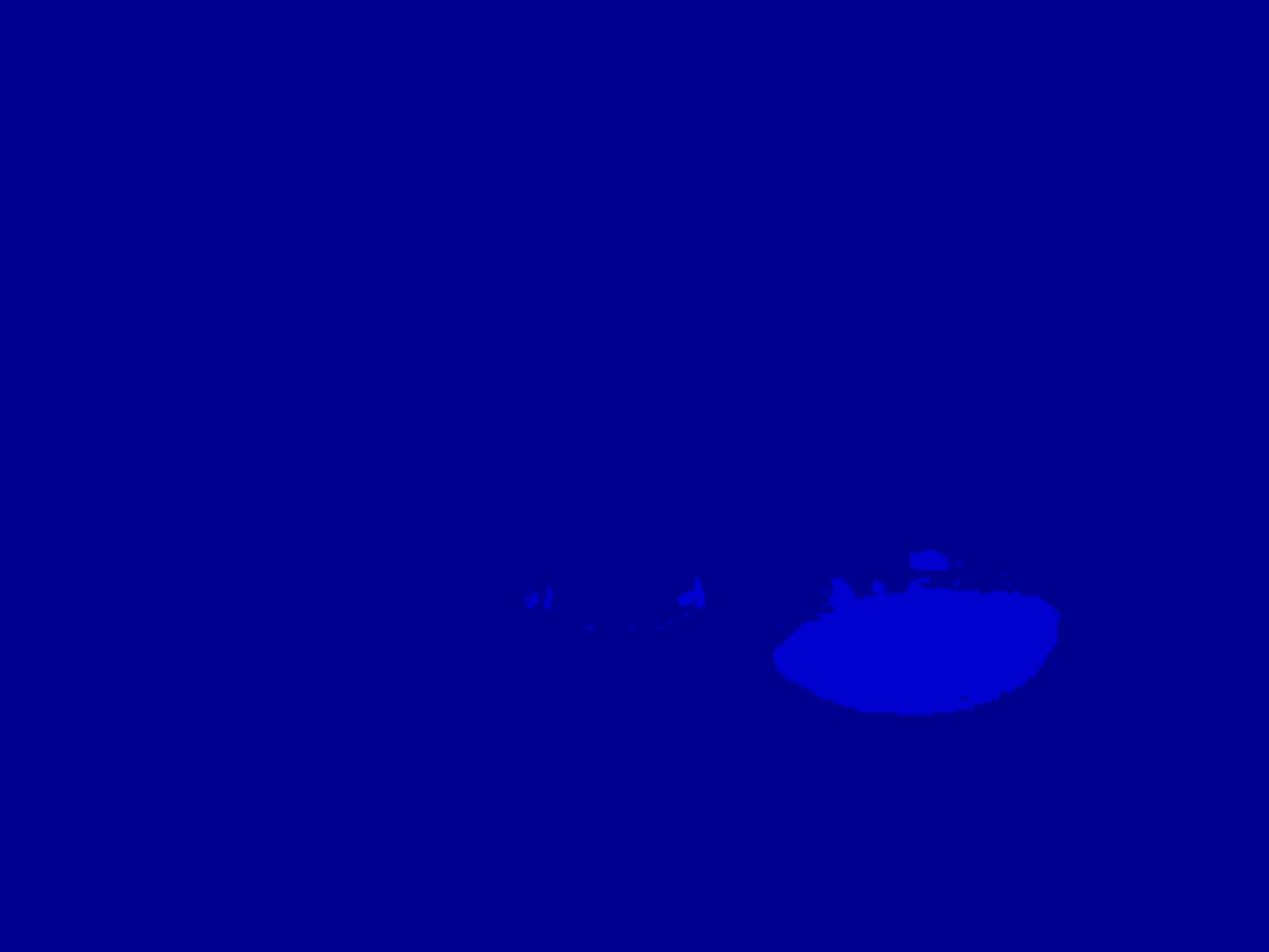}

\myfigurethreecol{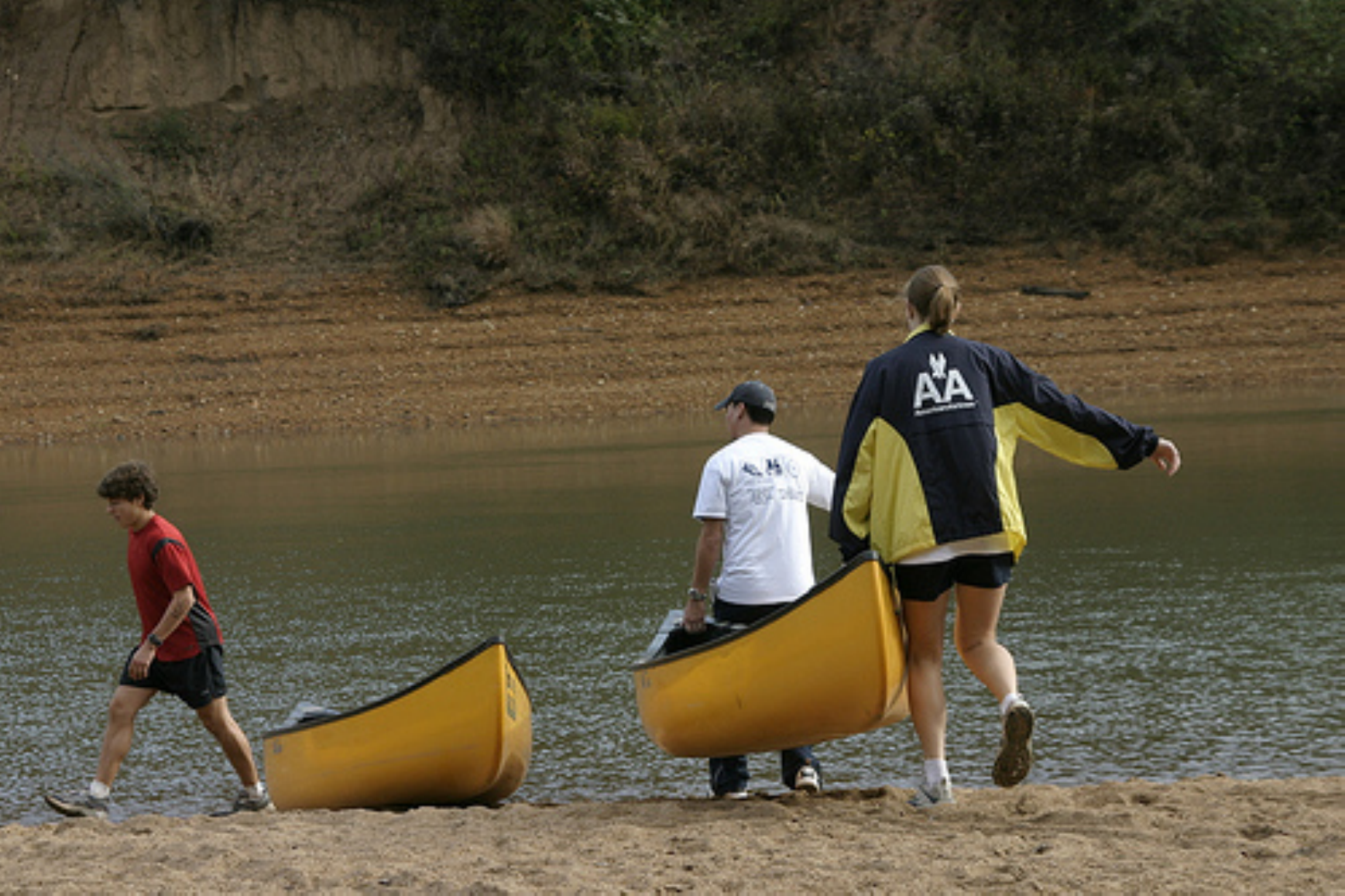}
\myfigurethreecol{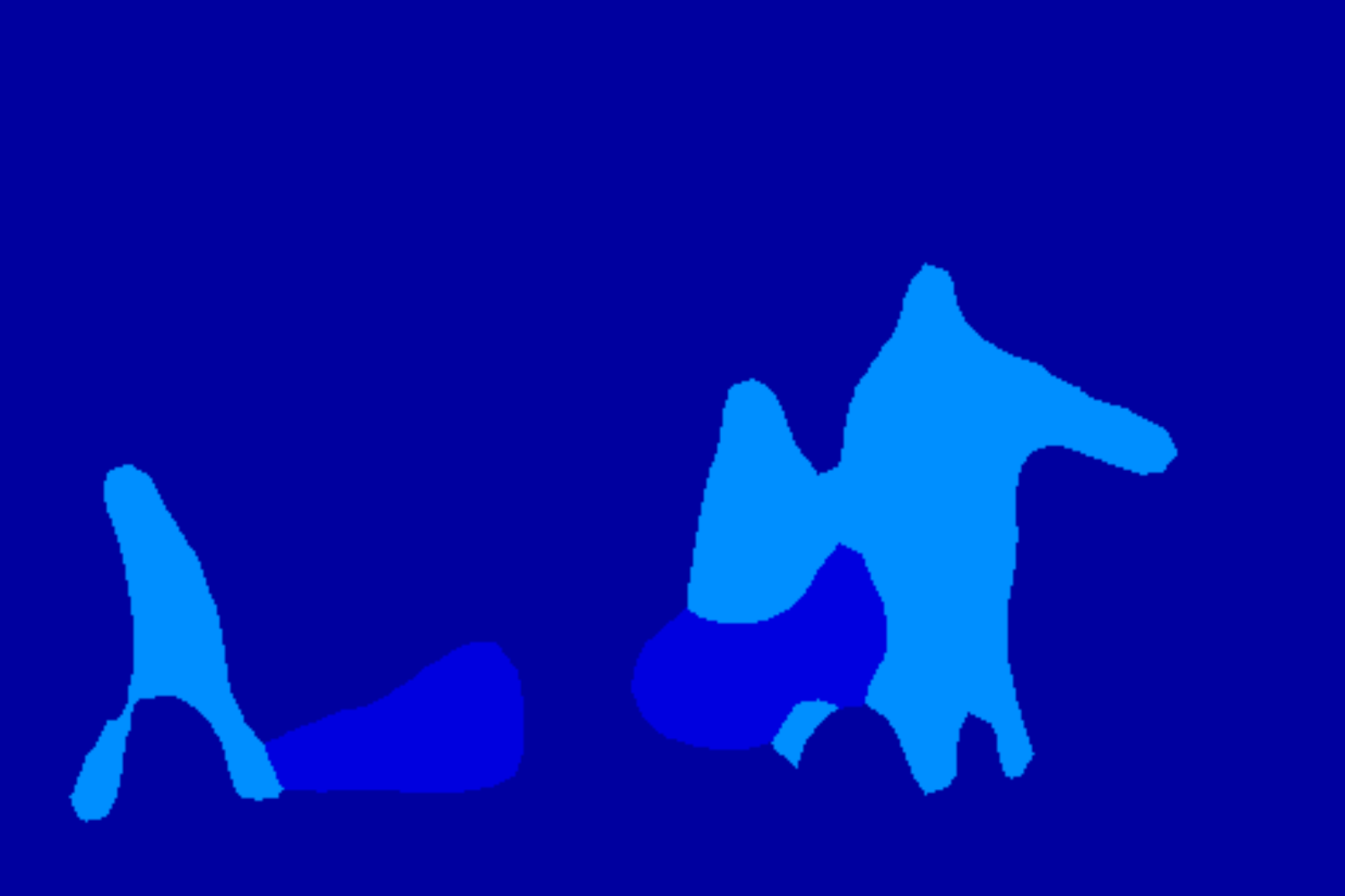}
\myfigurethreecol{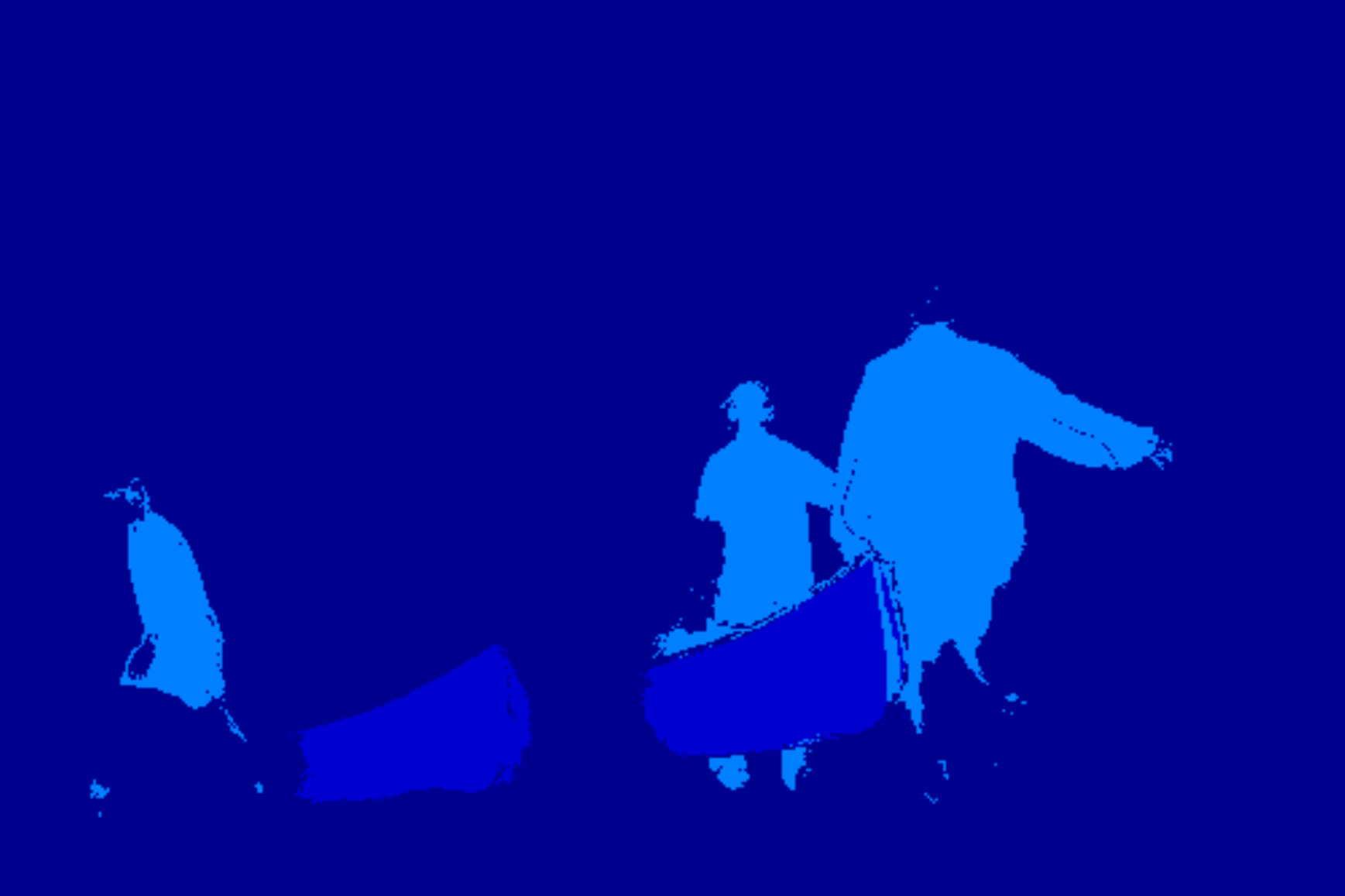}

%\myfigurethreecol{./paper_figures/intro/2008_000195_my_con_comp.pdf}
%\myfigurethreecol{./paper_figures/intro/2008_000195_my_seg.pdf}
%\myfigurethreecol{./paper_figures/intro/2008_000195_gt.pdf}

\captionsetup{labelformat=default}
\setcounter{figure}{0}
   %\caption{AA}
    %\caption{The first row represents an input image, predictions by FCN~\cite{long_shelhamer_fcn} and Dense-CRF~\cite{DBLP:journals/corr/ChenPKMY14} respectively. Note that segmentation produced by FCN is poorly localized around object boundaries. Also, observe that due to a poor pixel affinity measure, Dense-CRF produces spatially disjoint predictions. In comparison to these methods, our \textit{2FOR1 Net} combines semantic boundaries with the coarse segmentation to minimize the global loss that produces results more similar to the ground truth (bottom left image).}
    \caption{Examples illustrating shortcomings of prior semantic segmentation methods: the second column shows    
 results obtained with a FCN~\cite{long_shelhamer_fcn}, while the third column shows the output of a Dense-CRF applied to FCN predictions~\cite{NIPS2011_4296,DBLP:journals/corr/ChenPKMY14}. Segments produced by FCN are blob-like and are poorly localized around object boundaries. Dense-CRF produces spatially disjoint object segments due to the use of a color-based pixel affinity function that is unable to measure semantic similarity between pixels.\vspace{-0.5cm}}
    \label{intro_fig}
\end{figure}

%%%%%%%%% BODY TEXT
\section{Introduction}

The recent introduction of fully convolutional networks (FCNs)~\cite{long_shelhamer_fcn} has led to significant quantitative improvements on the task of semantic segmentation. However, despite their empirical success, FCNs suffer from some limitations. Large receptive fields in the convolutional layers and the presence of pooling layers lead to blurring and segmentation predictions at a significantly lower resolution than the original image. As a result, their predicted segments tend to be blobby and lack fine object boundary details. We report in Fig.~\ref{intro_fig} some examples illustrating typical poor localization of objects in the outputs of FCNs.

Recently, Chen at al.~\cite{DBLP:journals/corr/ChenPKMY14} addressed this issue by applying a Dense-CRF post-processing step~\cite{NIPS2011_4296} on top of coarse FCN segmentations. However, such an approach introduces several problems of its own. First, the Dense-CRF adds new parameters that are difficult to tune and integrate into the original network architecture. Additionally, most methods based on CRFs or MRFs use low-level pixel affinity functions, such as those based on color. These low-level affinities often fail to capture semantic relationships between objects and lead to poor segmentation results (see last column in Fig.~\ref{intro_fig}). 

%For instance, suppose we wanted to segment a person who is wearing black pants and a white shirt. Using a color based pixel affinity would produce a spatially disjoint prediction as illustrated in Fig.~\ref{intro_fig}. Instead, we would like to use pairwise pixel affinities that accurately captures the relationships between pixels for the task of semantic segmentation.

We propose to address these shortcomings by means of a Boundary Neural Field (BNF), an architecture that employs a single semantic segmentation FCN to predict semantic boundaries and then use them to produce semantic segmentation maps via a global optimization. We demonstrate that even though the semantic segmentation FCN has not been optimized to detect boundaries, it provides good features for  boundary detection. Specifically, the contributions of our work are as follows: 
\begin{itemize}
%\begin{enumerate}
\item We show that semantic boundaries can be expressed as a linear combination of interpolated convolutional feature maps inside an FCN. We introduce a boundary detection method that exploits this intuition to predict object boundaries with accuracy superior to the state-the-of-art.
\item We demonstrate that boundary-based pixel affinities are better suited for semantic segmentation than the commonly used color affinity functions.
\item Finally, we introduce a new global energy that decomposes semantic segmentation into multiple binary problems and relaxes the integrality constraint. We show that minimizing our proposed energy yields better qualitative and quantitative results relative to traditional globalization models such as MRFs or CRFs.
\end{itemize}
%\end{enumerate}

%Thus, our contributions include a method that combines boundary detection and semantic segmentation and achieves excellent results in both tasks, a global loss that models semantic segmentation in a novel way and can be integrated into the overall network architecture, and our proposed boundary based pixel affinity that leads to better semantic segmentation results.

\begin{figure}
\begin{center}
   \includegraphics[width=1\linewidth]{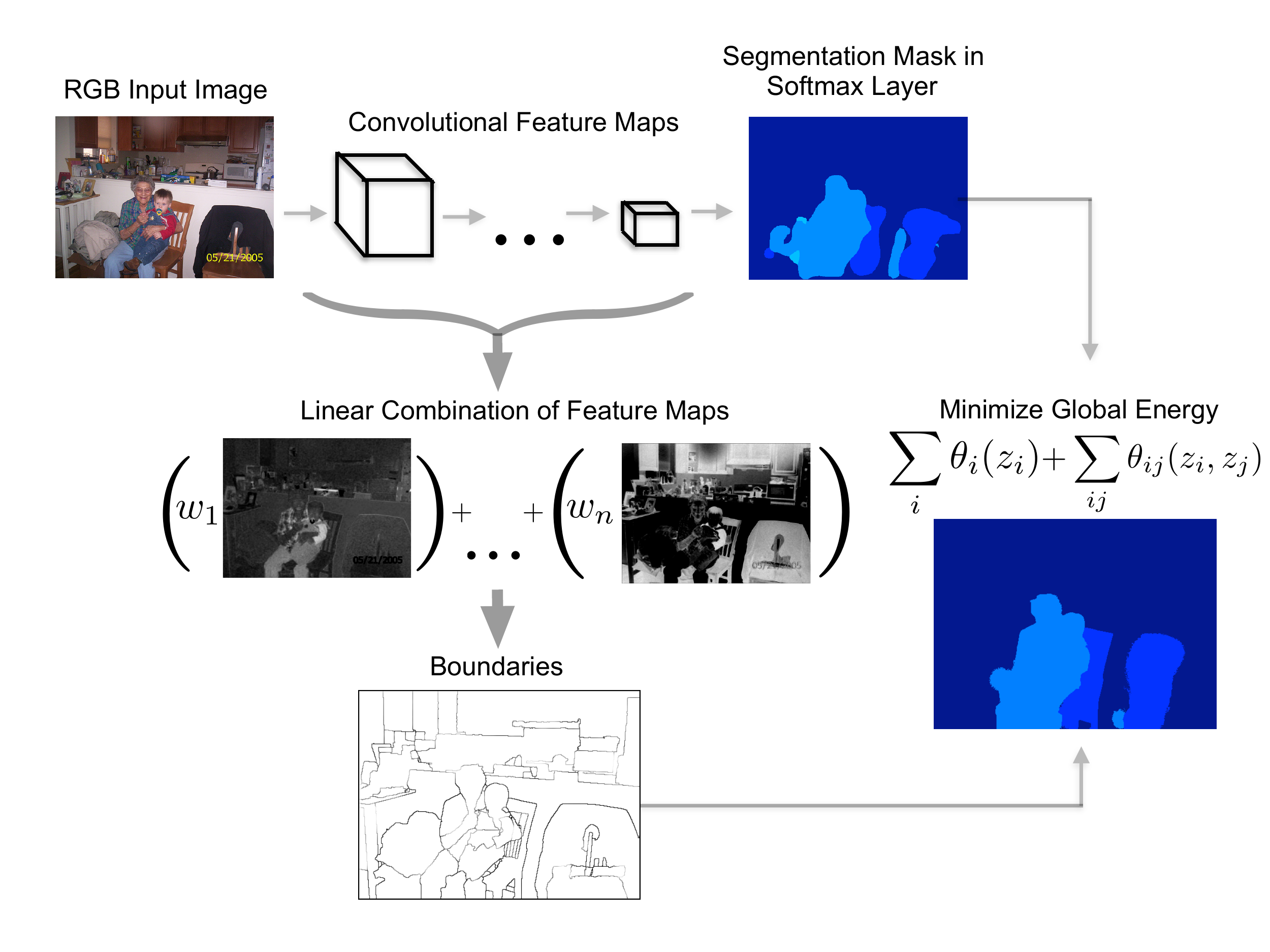}
\end{center}
   \caption{The architecture of our system (best viewed in color). We employ a semantic segmentation FCN~\cite{DBLP:journals/corr/ChenPKMY14} for two purposes: 1) to obtain semantic segmentation unaries for our global energy; 2) to compute object boundaries. Specifically, we define semantic boundaries as a linear combination of these feature maps (with a sigmoid function applied on top of the sum) and learn individual weights corresponding to each convolutional feature map. We integrate this boundary information in the form of pairwise potentials (pixel affinities) for our energy model.}%\vspace{-0.4cm}}
\label{fig:arch}
\end{figure}

\section{Related Work}

\textbf{Boundary Detection.} Spectral methods comprise one of the most prominent categories for boundary detection. In a typical spectral framework, one formulates a generalized eigenvalue system to solve a low-level pixel grouping problem. The resulting eigenvectors are then used to predict the boundaries. Some of the most notable approaches in this genre are MCG~\cite{cArbelaez14}, gPb~\cite{Arbelaez:2011:CDH:1963053.1963088}, PMI~\cite{crisp_boundaries}, and Normalized Cuts~\cite{Shi97normalizedcuts}. A weakness of spectral approaches is that they tend to be slow as they perform a global inference over the entire image.

To address this issue, recent approaches cast boundary detection as a classification problem and predict the boundaries in a local manner with high efficiency.  The most notable examples in this genre include sketch tokens (ST)~\cite{LimCVPR13SketchTokens} and structured edges (SE)~\cite{Dollar2015PAMI}, which employ fast random forests. However, many of these methods are based on hand-constructed features, which are difficult to tune.

%and sparse code gradients (SCG)~\cite{ren_nips12}.
    %Whereas the first two methods utilize random forests to classify boundaries, the latter method employs an SVM classifier.

The issue of hand-constructed features have been recently addressed by several approaches based on deep learning, such as $N^4$ fields~\cite{DBLP:journals/corr/GaninL14}, DeepNet~\cite{kivinen2014visual}, DeepContour~\cite{Shen_2015_CVPR}, DeepEdge~\cite{gberta_2015_CVPR}, \HfL~\cite{gberta_2015_ICCV} and HED~\cite{DBLP:journals/corr/XieT15}. All of these methods use CNNs in some way to predict the boundaries. Whereas DeepNet and DeepContour optimize ordinary CNNs to a boundary based optimization criterion from scratch, DeepEdge and \HfL employ pretrained models to compute boundaries. The most recent of these methods is HED~\cite{DBLP:journals/corr/XieT15}, which shows the benefit of deeply supervised learning for boundary detection.

%\LT{The 3 paragraphs above read a bit like a laundry list.}

In comparison to prior deep learning approaches, our method offers several contributions. First, we exploit the inherent relationship between boundary detection and semantic segmentation to predict semantic boundaries. Specifically, we show that even though the semantic FCN has not been explicitly trained to predict boundaries, the convolutional filters inside the FCN provide good features for boundary detection. Additionally, unlike DeepEdge~\cite{gberta_2015_CVPR} and \HfL~\cite{gberta_2015_ICCV}, our method does not require a pre-processing step to select candidate contour points, as we predict boundaries on all pixels in the image. We demonstrate that our approach allows us to achieve state-of-the-art boundary detection results according to both F-score and Average Precision metrics. Additionally, due to the semantic nature of our boundaries, we can successfully use them as pairwise potentials for semantic segmentation in order to improve object localization and recover fine structural details, typically lost by pure FCN-based approaches.

\textbf{Semantic Segmentation.} We can group most  semantic segmentation methods into three broad categories. The first category can be described as ``two-stage'' approaches, where an image is first segmented and then each segment is classified as belonging to a certain object class. Some of the most notable methods that belong to this genre include~\cite{DBLP:journals/corr/MostajabiYS14,Carreira:2012:SSS:2403272.2403306, gupta14rcnndepth,BharathECCV2014}.

The primary weakness of the above methods is that they are unable to recover from errors made by the segmentation algorithm. Several recent papers~\cite{DBLP:journals/corr/HariharanAGM14a,farabet-pami-13} address this issue by proposing to use deep per-pixel CNN features and then classify each pixel as belonging to a certain class. While these approaches partially address the incorrect segmentation problem, they perform predictions independently on each pixel. This leads to extremely local predictions, where the relationships between pixels are not exploited in any way, and thus the resulting segmentations may be spatially disjoint. 

%can still be viewed as a two-stage approach where segmentation mechanism is employed at a later stage to refine the predictions.

%\LT{this is a bit weak as a critique}.

The third and final group of semantic segmentation methods can be viewed as front-to-end schemes where segmentation maps are predicted directly from raw pixels without any intermediate steps. One of the earliest examples of such methods is the FCN introduced in~\cite{long_shelhamer_fcn}. This approach gave rise to a number of subsequent related approaches which have improved various aspects of the original semantic segmentation~\cite{DBLP:journals/corr/ChenPKMY14,crfasrnn_iccv2015,DBLP:journals/corr/DaiH015,hong2015decoupled,DBLP:journals/corr/LinSRH15}.
There have also been attempts at integrating the CRF mechanism into the network architecture~\cite{DBLP:journals/corr/ChenPKMY14,crfasrnn_iccv2015}. Finally, it has been shown that semantic segmentation can also be improved using additional training data in the form of bounding boxes~\cite{DBLP:journals/corr/DaiH015}. 

%Conceptually, the most similar to our work is DSM~\cite{DBLP:journals/corr/LinSRH15}, which explicitly tries to learn unary and pairwise potentials and then globalize them using CRF loss. However, unlike our method which encodes pairwise potentials in a form of semantic boundaries, DSM learns pairwise potentials that encode spatial ``below/above'', ``surrounding'', type of relationships. Additionally, DSM uses two separate networks to learn unary and pairwise potentials, which essentially doubles the number of parameters, which is large even on a single network ($\approx 15M$). Contrary to DSM, we can learn pairwise potentials in a form of semantic boundaries after adding only $\approx 5K $ parameters, which is about $3$ orders of magnitude less. Finally, unlike DSM, the solution to our global loss can be obtained via a closed-form solution, which makes inference efficient.

%($\approx 15M$)
%( $\approx 15M$) 

Our BNF offers several contributions over prior work. To the best of our knowledge,  we are the first to present a model that exploits the relationship between boundary detection and semantic segmentation \textbf{within a FCN framework}. We introduce pairwise pixel affinities computed from semantic boundaries inside an FCN, and use these boundaries to predict the segmentations in a global fashion. Unlike~\cite{DBLP:journals/corr/LinSRH15}, which requires a large number of additional parameters to learn for the pairwise potentials, our global model only needs $\approx 5K$ extra parameters, which is about $3$ orders of magnitudes less than the number of parameters in a typical deep convolutional network (e.g. VGG~\cite{Simonyan14c}). We empirically show that our proposed boundary-based affinities are better suited for semantic segmentation than color-based affinities. Additionally, unlike in~\cite{DBLP:journals/corr/ChenPKMY14,crfasrnn_iccv2015,DBLP:journals/corr/LinSRH15}, the solution to our proposed global energy can be obtained in closed-form, which makes global inference easier. Finally we demonstrate that our method produces better results than traditional globalization models such as CRFs or MRFs.

%\section{2FOR1 Net}
\section{Boundary Neural Fields}

%describe general idea

In this section, we describe Boundary Neural Fields. %Similar to prior semantic segmentation approaches, we employ FCNs~\cite{long_shelhamer_fcn}. 
Similarly to traditional globalization methods, Boundary Neural Fields are defined by an energy including unary and pairwise potentials.  Minimization of the global energy yields the semantic segmentation. BNFs build both unary and pairwise potentials from the input RGB image and then combine them in a global manner. More precisely, the coarse segmentations predicted by a semantic FCN are used to define the unary potentials of our BNF. Next, we show that the convolutional feature maps of the FCN can be used to accurately predict semantic boundaries. These boundaries are then used to build pairwise pixel affinities, which are used as pairwise potentials by the BNF. Finally, we introduce a global energy function, which minimizes the energy corresponding to the unary and pairwise terms and improves the initial FCN segmentation. The detailed illustration of our architecture is presented in Figure~\ref{fig:arch}. We now explain each of these steps in more detail.

\subsection{FCN Unary Potentials}

%mention that we can  use any state-of-the-art network for our unary potential rpedictor

%\LT{do we use the CRF component of it? The word CRF here is confusing.} 

%we use a deeplab-crf which is fully convolutional adaptation of VGG network

To predict semantic unary potentials we employ the DeepLab model~\cite{DBLP:journals/corr/ChenPKMY14}, which is a fully convolutional adaptation of the VGG network~\cite{Simonyan14c}. The FCN consists of $16$ convolutional layers and $3$ fully convolutional layers. There are more recent FCN-based methods that have demonstrated even better semantic segmentation results~\cite{DBLP:journals/corr/DaiH015,crfasrnn_iccv2015,hong2015decoupled,DBLP:journals/corr/LinSRH15}. Although these more advanced architectures could be integrated into our framework to improve our unary potentials, in this work we focus on two aspects orthogonal to this prior work: 1) demonstrating that our boundary-based affinity function is better suited for semantic segmentation than the common color-based affinities and 2) showing that our proposed global energy achieves better qualitative and quantitative semantic segmentation results in comparison to prior globalization models.

%We note that  pretrained DeepLab-CRF model, which we use to initialize our network, has been trained on a large Microsoft-COCO dataset~\cite{502}. Therefore, it is enough to fine-tune our network for $2000$ iterations.

\subsection{Boundary Pairwise Potentials}

%visualize magnitudes of filters
%visualize highest ranked filters
%visualize outputs  obtained from this second branch
%mention NMS post-processing

In this section, we describe our approach for building pairwise pixel affinities using semantic boundaries. The basic idea behind our boundary detection approach is to express semantic boundaries as a function of convolutional feature maps inside the FCN. Due to the close relationship between the tasks of semantic segmentation and boundary detection, we hypothesize that convolutional feature maps from the semantic segmentation FCN can be employed as features for boundary detection. 

%Due to large receptive fields and the presence of pooling layers, both of which cause blurring and FCN predictions at a lower resolution, FCN cannot effectively use the information from all the convolutional feature maps. However, we demonstrate that  even though the FCN has not been explicitly trained to predict boundaries , these features maps inside FCN provide good features for the boundary detection task.

%N fcn these feature maps cannot be used effectively due to  large-receptive fields and pooling layers, but they contain boundary information

%At first, this may seem counter-intuitive, because we already know that FCN produces segmentation that does not capture object boundaries well. For this reason, one may hypothesize that the FCN simply did not learn relevant boundary information. However, we demonstrate that FCN does contain accurate boundary information and we simply need to extract these boundaries out of the FCN.

%We first explain how to predict semantic boundaries using the convolutional feature maps. Then, we demonstrate how to use these boundaries to encode pairwise relationships between pixels.

\subsubsection{Learning to Predict Semantic Boundaries.}  We propose to express semantic boundaries as a linear combination of interpolated FCN feature maps with a non-linear function applied on top of this sum. We note that interpolation of feature maps has been successfully used  in prior work (see e.g.~\cite{DBLP:journals/corr/HariharanAGM14a}) in order to obtain dense pixel-level features from the low-resolution outputs of deep convolutional layers. Here we adopt interpolation to produce pixel-level boundary predictions. There are several advantages to our proposed formulation. First, because we express boundaries as a linear combination of feature maps, we only need to learn a small number of parameters, corresponding to the individual weight values of each feature map in the FCN. This amounts to  $\approx 5K$ learning parameters, which is much smaller than the number of parameters in the entire network ($\approx 15M$). In comparison, DeepEdge~\cite{gberta_2015_CVPR} and HFL~\cite{gberta_2015_ICCV} need 17M and 6M {\em additional} parameters to predict boundaries.

%Despite using only a small number of extra parameters, our method achieves better accuracy on BSDS compared to DeepEdge~\cite{gberta_2015_CVPR}, HFL~\cite{gberta_2015_ICCV}, and HED~\cite{DBLP:journals/corr/XieT15} .

%This implies that we can perform boundary detection and then use these boundaries to improve semantic segmentation with little added training cost.

Furthermore, expressing semantic boundaries as a linear combination of FCN feature maps allows us to efficiently predict boundary probabilities for all pixels in the image (we resize the FCN feature maps to the original image dimensions). This eliminates the need to select candidate boundary points in a pre-processing stage, which was instead required in prior boundary detection work~\cite{gberta_2015_CVPR,gberta_2015_ICCV}.

Our boundary prediction pipeline can be described as follows. First we use use SBD segmentations~\cite{BharathICCV2011} to optimize our FCN for semantic segmentation task. We then treat FCN convolutional maps as features for the boundary detection task and use the boundary annotations from BSDS 500 dataset~\cite{MartinFTM01} to learn the weights for each feature map. BSDS 500 dataset contains $200$ training, $100$ validation, $200$ testing images, and ground truth annotations by $5$ human labelers for each of these images. 

To learn the weights corresponding to each convolutional feature map we first sample $80K$ points from the dataset. We define the target labels for each point as the fraction of human annotators agreeing on that point being a boundary. To fix the issue of label imbalance (there are many more non-boundaries than boundaries), we divide the label space into four quartiles, and select an equal number of samples for each quartile to balance the training dataset. Given these sampled points, we then define our features as the values in the interpolated convolutional feature maps corresponding to these points. To predict semantic boundaries we weigh each convolutional feature map by its weight, sum them up and apply a sigmoid function on top of it. We obtain the weights corresponding to each convolutional feature map by minimizing the cross-entropy loss using a stochastic batch gradient descent for $50$ epochs. To obtain crisper boundaries at test-time we  post-process the boundary probabilities using non-maximum suppression.

\begin{figure}
\centering

\myfigurethreecol{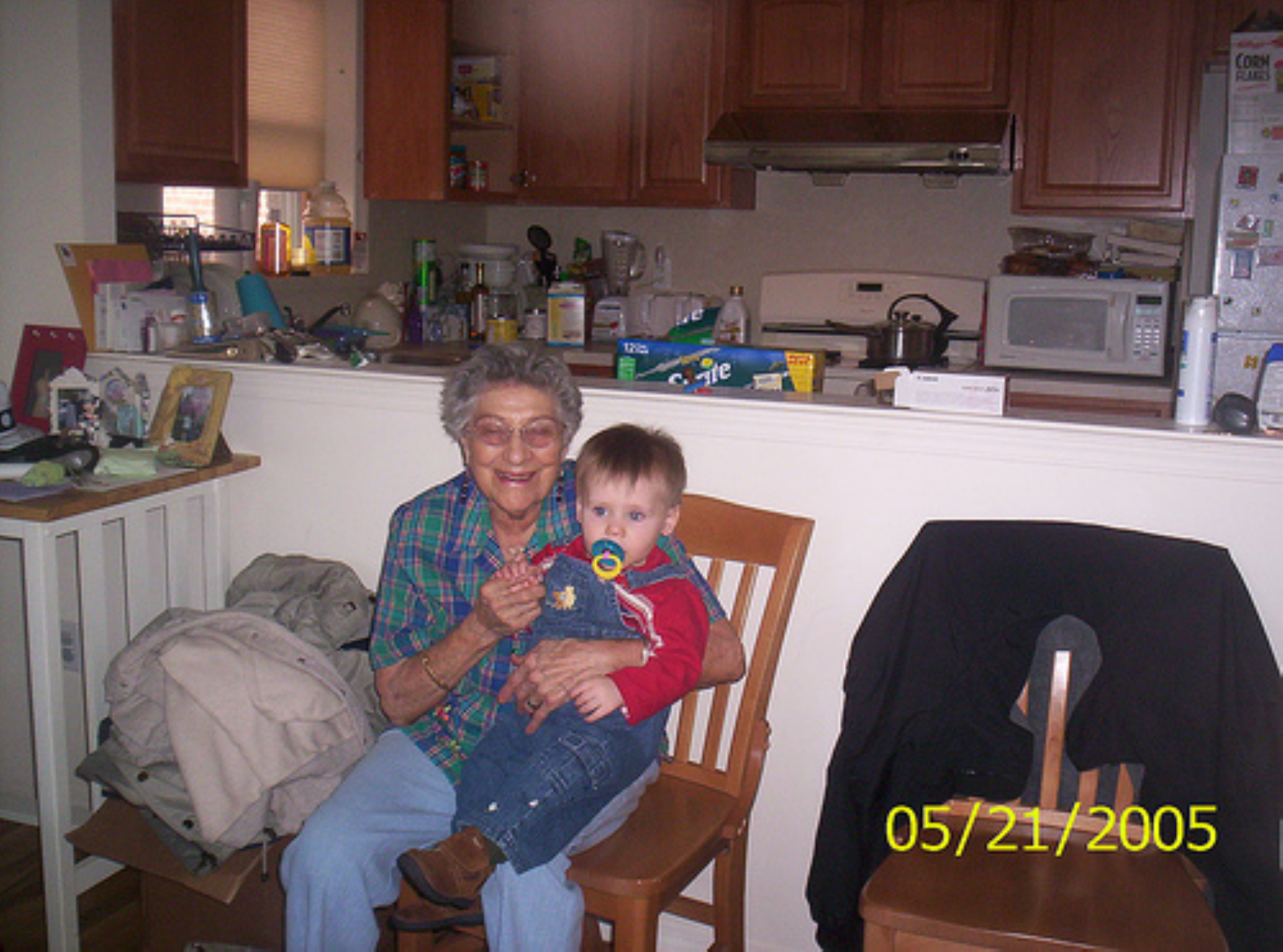}
\myfigurethreecol{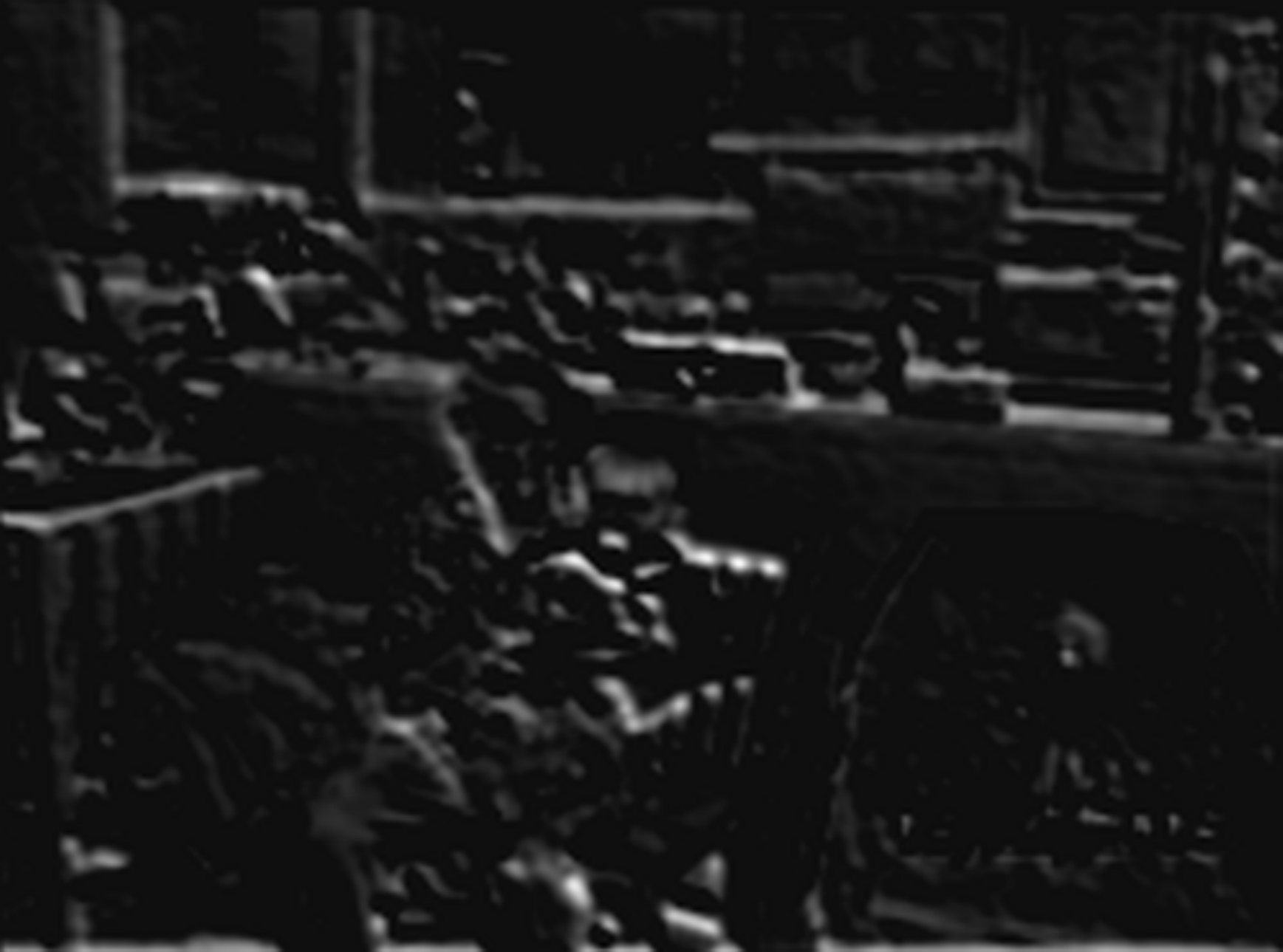}
\myfigurethreecol{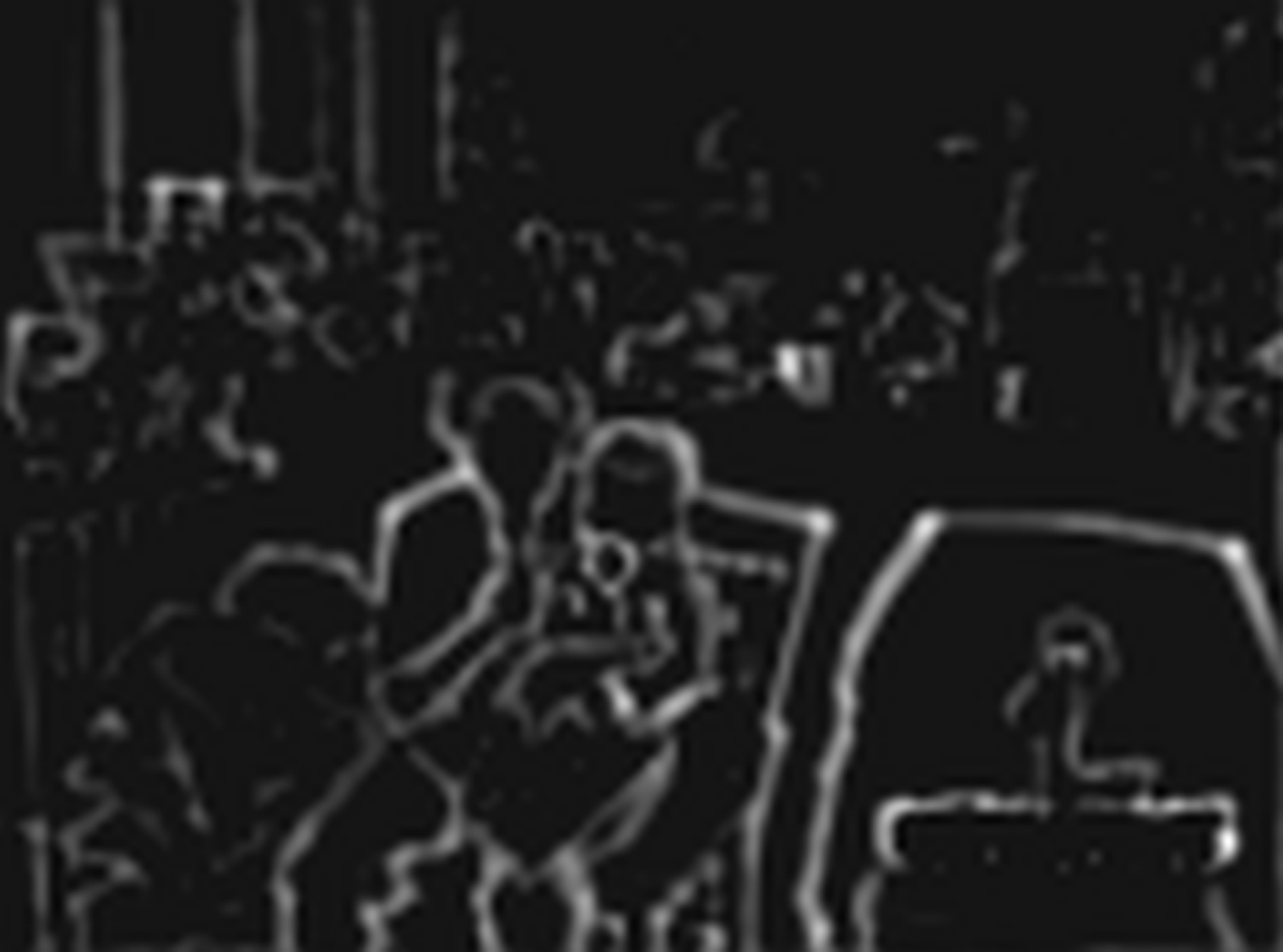}

\myfigurethreecol{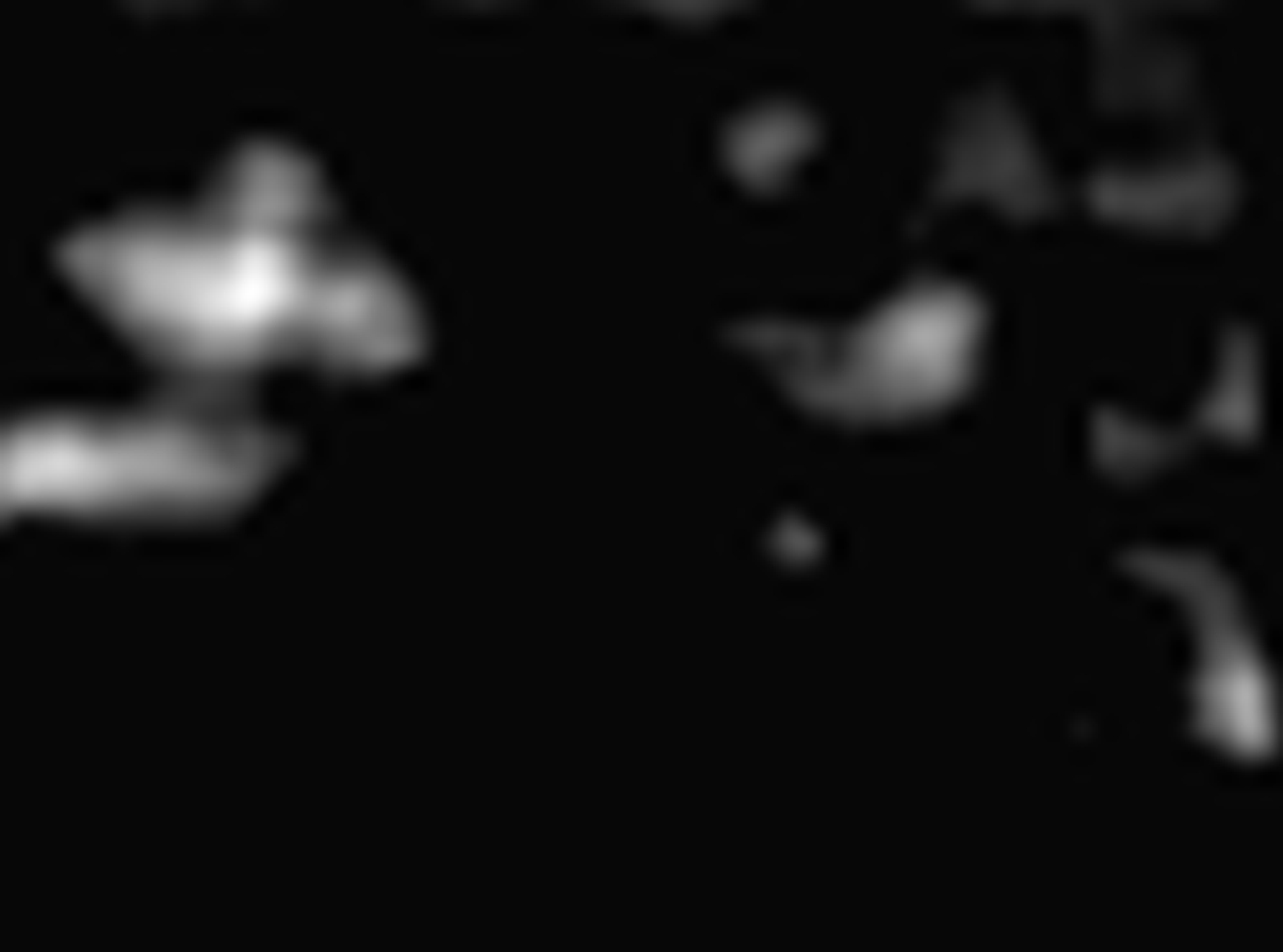}
\myfigurethreecol{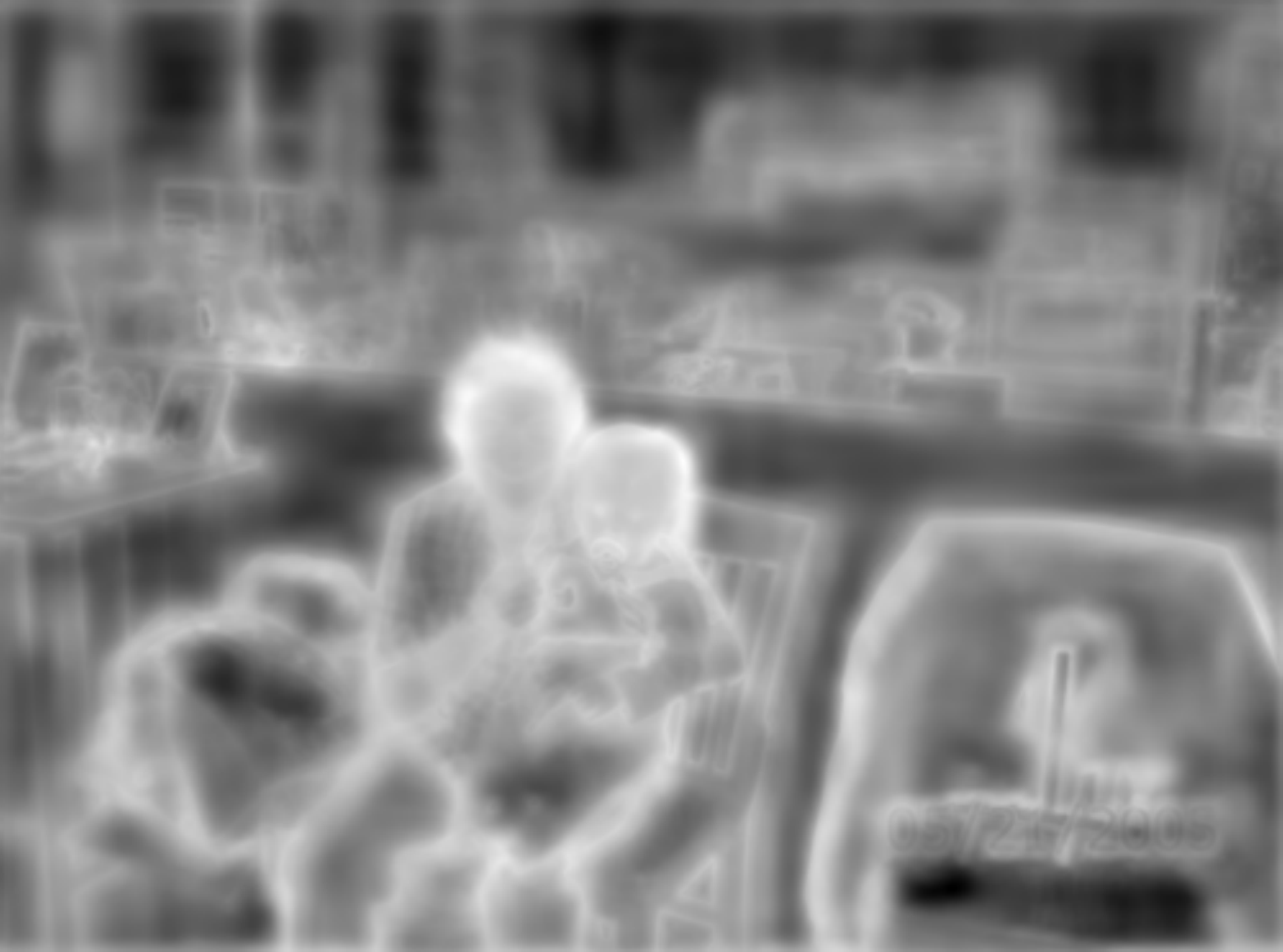}
\myfigurethreecol{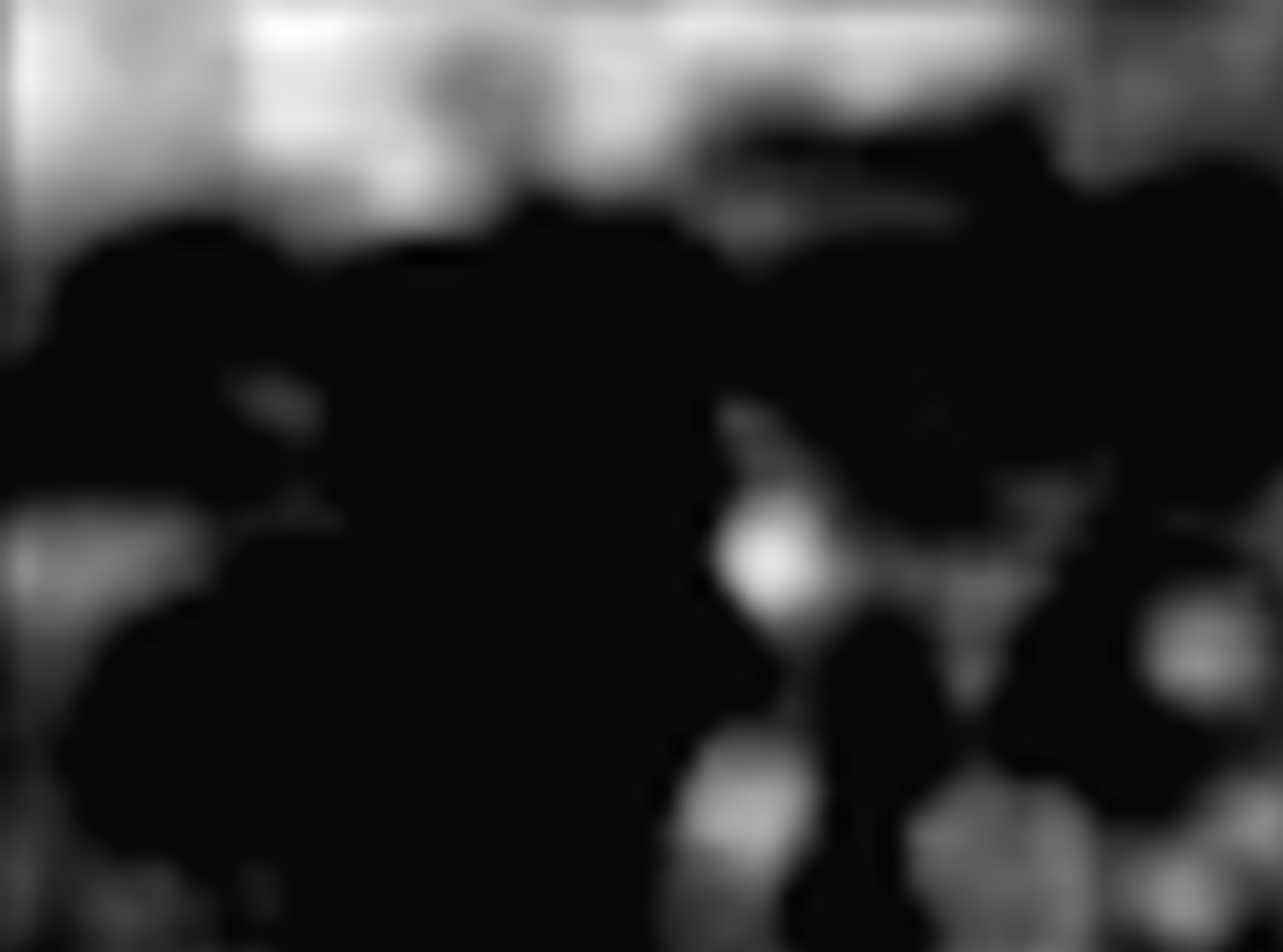}

\captionsetup{labelformat=default}
%\setcounter{figure}{1}
   %\caption{AA}
    \caption{An input image and convolutional feature maps corresponding to the largest weight magnitude values. Intuitively these are the feature maps that contribute most heavily to the task of boundary detection.}
    \label{conv_maps}
\end{figure}

To give some intuition on how FCN feature maps contribute to boundary detection, in Fig.~\ref{conv_maps} we visualize the feature maps corresponding to the highest weight magnitudes. It is clear that many of these maps contain highly localized boundary information.

%To give some intuition on how FCN feature maps contribute to boundary detection, in Fig.~\ref{fig:weights} we visualize the weight magnitudes corresponding to each layer (from the earliest to the deepest) in our FCN. Note that the weight magnitudes of the deepest layers have slightly larger values. This implies that semantic segmentation cues are beneficial for boundary detection. Additionally, in Fig.~\ref{conv_maps} we visualize the feature maps corresponding to the highest weight magnitudes. It is clear that many of these maps contain highly localized boundary information.

\captionsetup{labelformat=empty}

\captionsetup{labelformat=default}

\captionsetup{labelformat=default}

  \begin{table}
    \begin{center}
    \begin{tabular}{ | c | c | c | c |}
    \hline
    Method & ODS & OIS & AP \\ \hline\hline
	SCG~\cite{ren_nips12} & 0.739 & 0.758 & 0.773 \\ \hline
	SE~\cite{Dollar2015PAMI}  & 0.746 & 0.767 & 0.803\\ \hline
	MCG~\cite{cArbelaez14} & 0.747 & 0.779 & 0.759\\ \hline
	$N^4$-fields~\cite{DBLP:journals/corr/GaninL14} & 0.753 & 0.769 & 0.784\\ \hline
	DeepEdge~\cite{gberta_2015_CVPR}  & 0.753 & 0.772 & 0.807\\ \hline
	DeepContour~\cite{Shen_2015_CVPR} & 0.756 & 0.773 & 0.797\\ \hline
	\HfL~\cite{gberta_2015_ICCV}  & 0.767 & 0.788 & 0.795\\ \hline
	HED~\cite{DBLP:journals/corr/XieT15} & 0.782 & 0.804 & 0.833\\ \hline
	%\bf MNet & 0.779 & 0.792 & 0.831\\  \hline
	\bf BNF & \bf 0.788 & \bf 0.807 & \bf 0.851\\  
	
    \hline
    \end{tabular}
    \end{center}
    \caption{Boundary detection results on BSDS500 benchmark. Our proposed method outperforms all prior algorithms according to all three evaluation metrics.}
    \label{any_bsds}
   \end{table}

%The most likely reason why object-level boundaries are not preserved in the final segmentation maps, is the fact that FCN's loss for semantic segmentation is not modeled to preserve the boundaries. Designing an optimization criterion that leads to an accurate segmentation and also preserved object boundaries would be one way to solve this issue. However, coming up with a loss formulation that captures these two characteristics is a non-trivial task so we resort to an alternative method to extract semantic boundaries out of FCN's feature maps. 

%We note that feature interpolation in deep layers have been succesfully applied for object segmentation and detection in~\cite{DBLP:journals/corr/SermanetEZMFL13, DBLP:journals/corr/HariharanAGM14a} and for the the boundary detection in~\cite{gberta_2015_ICCV}. However, unlike~\cite{gberta_2015_ICCV}, we predict boundaries using full convolutional feature maps rather than individual points. That is 

\textbf{Boundary Detection Results}
Before discussing how boundary information is integrated in our energy for semantic segmentation, here we present experimental results assessing the accuracy of our boundary detection scheme. We tested our boundary detector on the BSDS500 dataset~\cite{MartinFTM01}, which is the standard benchmark for boundary detection. The quality of the predicted boundaries is evaluated using three standard measures: fixed contour threshold (ODS), per-image best threshold (OIS), and average precision (AP).

%We compare our approach to the state-of-the-art methods by matching each of the predicted boundary pixels with the ground truth boundaries that were annotated by {\em any} of the human annotators. This is a standard way to evaluate the quality of our boundaries. 

In Table~\ref{any_bsds} we show that our algorithm outperforms all prior methods according to both F-score measures and the Average Precision metric.  In Fig.~\ref{raw_con}, we also visualize our predicted boundaries. The second column shows the pixel-level softmax output computed from the linear combination of feature maps, while the third column depicts our final boundaries after applying a non-maximum suppression post-processing step. 

We note that our predicted boundaries achieve high-confidence predictions around objects. This is important as we employ these boundaries to improve semantic segmentation results, as discussed in the next subsection.

%Even though this is expected since the FCN was optimized to a semantic segmentation criterion, the semantic nature of our boundaries 

%Finally, Fig.~\ref{raw_con},  depicts the output of a weighed linear combination of all feature maps with a sigmoid function applied on top of it. From the latter visualization, we can see that using the feature maps from FCN we can predict semantic boundaries pretty accurately. In order to get crisper boundaries from these maps, we post-process them using non-maximum suppression.

\begin{figure}
\centering

\myfigurethreecol{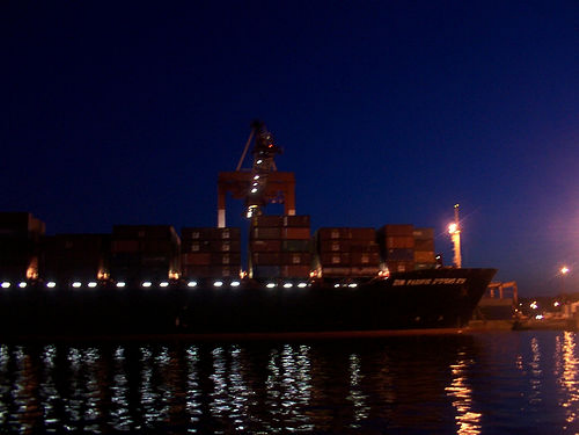}
\myfigurethreecol{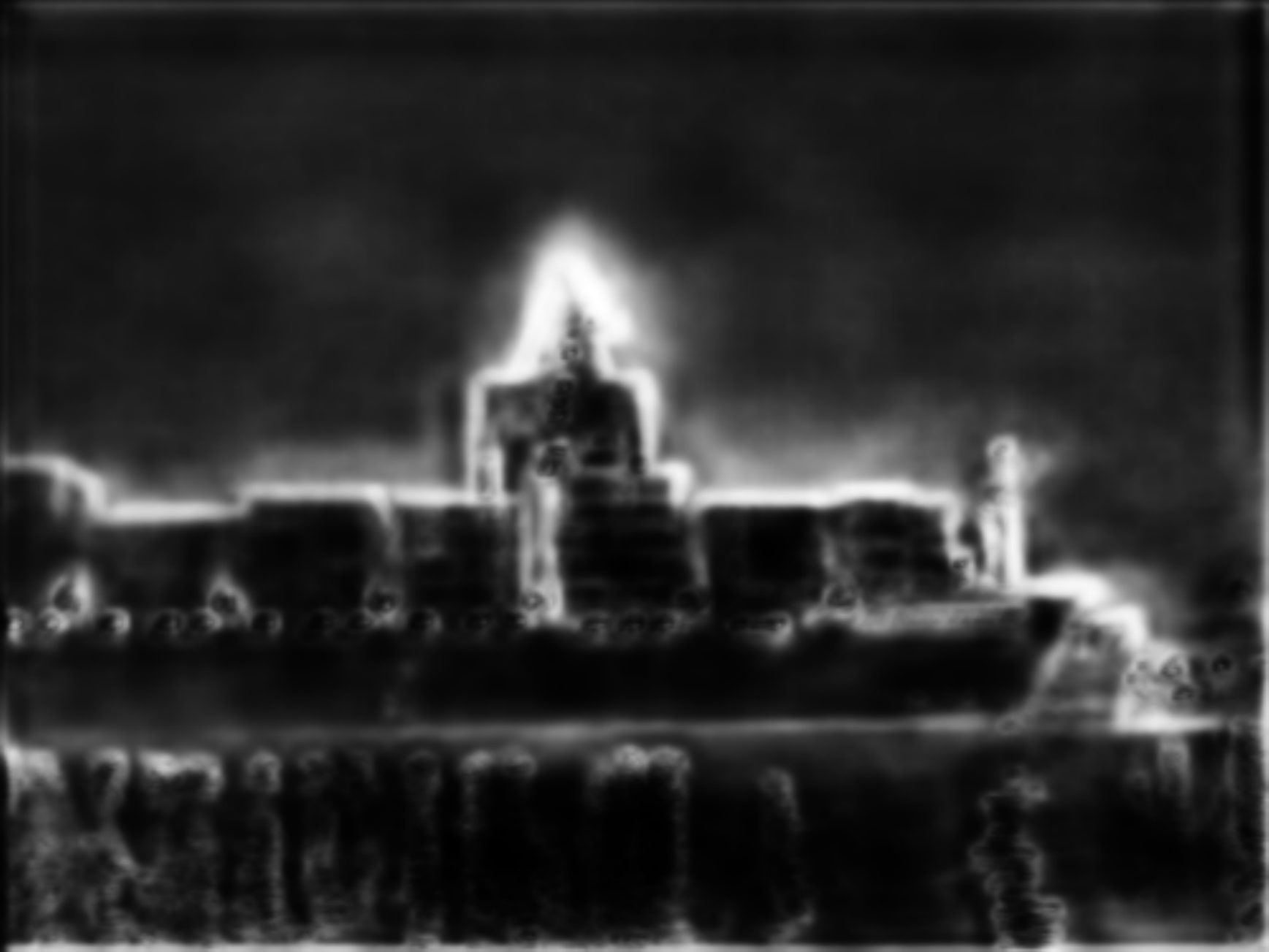}
\myfigurethreecol{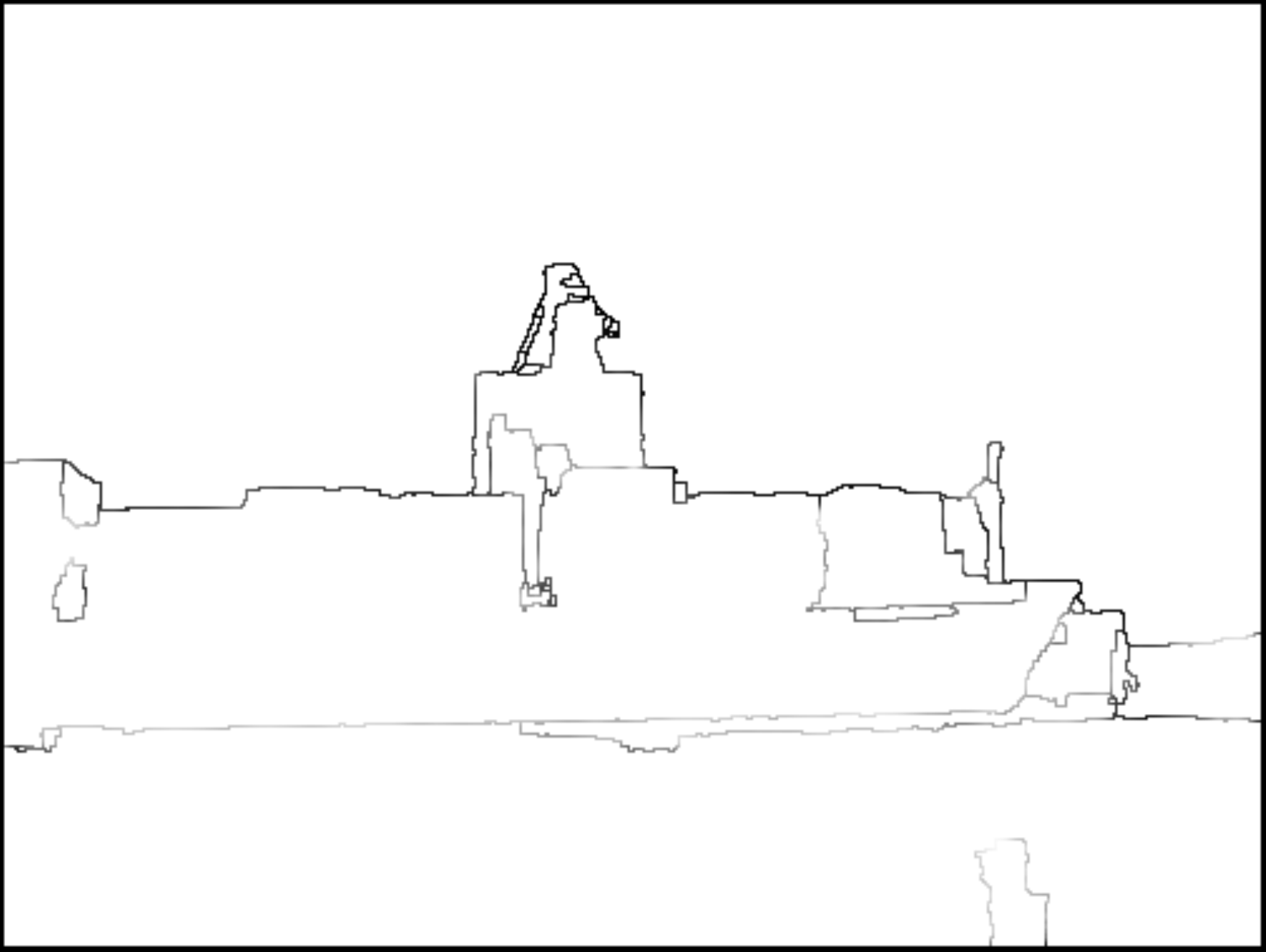}

%\myfigurethreecol{./paper_figures/raw_boundaries/2008_000123.pdf}
%\myfigurethreecol{./paper_figures/raw_boundaries/2008_000123_raw.pdf}
%\myfigurethreecol{./paper_figures/raw_boundaries/2008_000123_con.pdf}

\myfigurethreecol{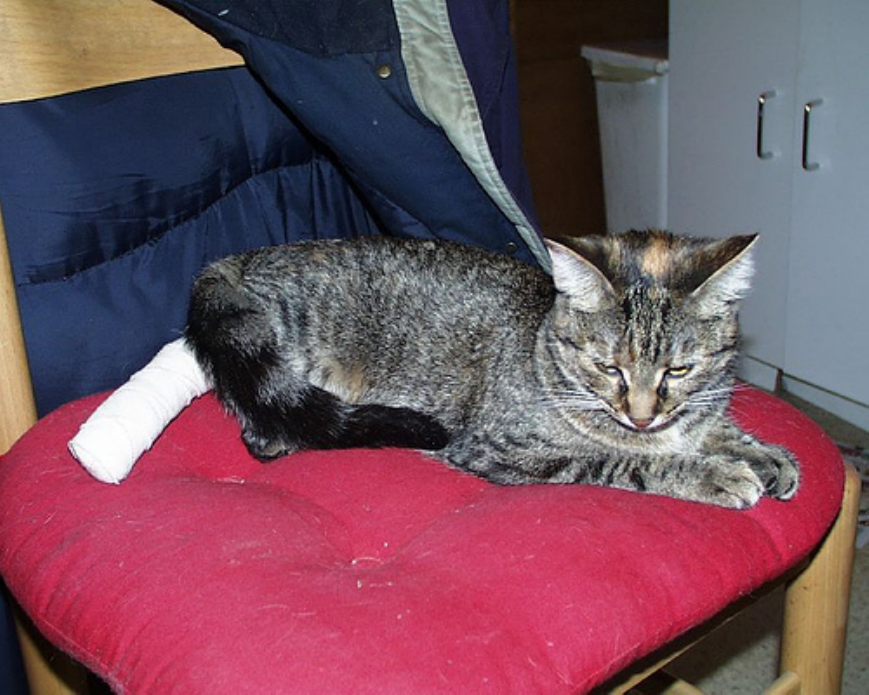}
\myfigurethreecol{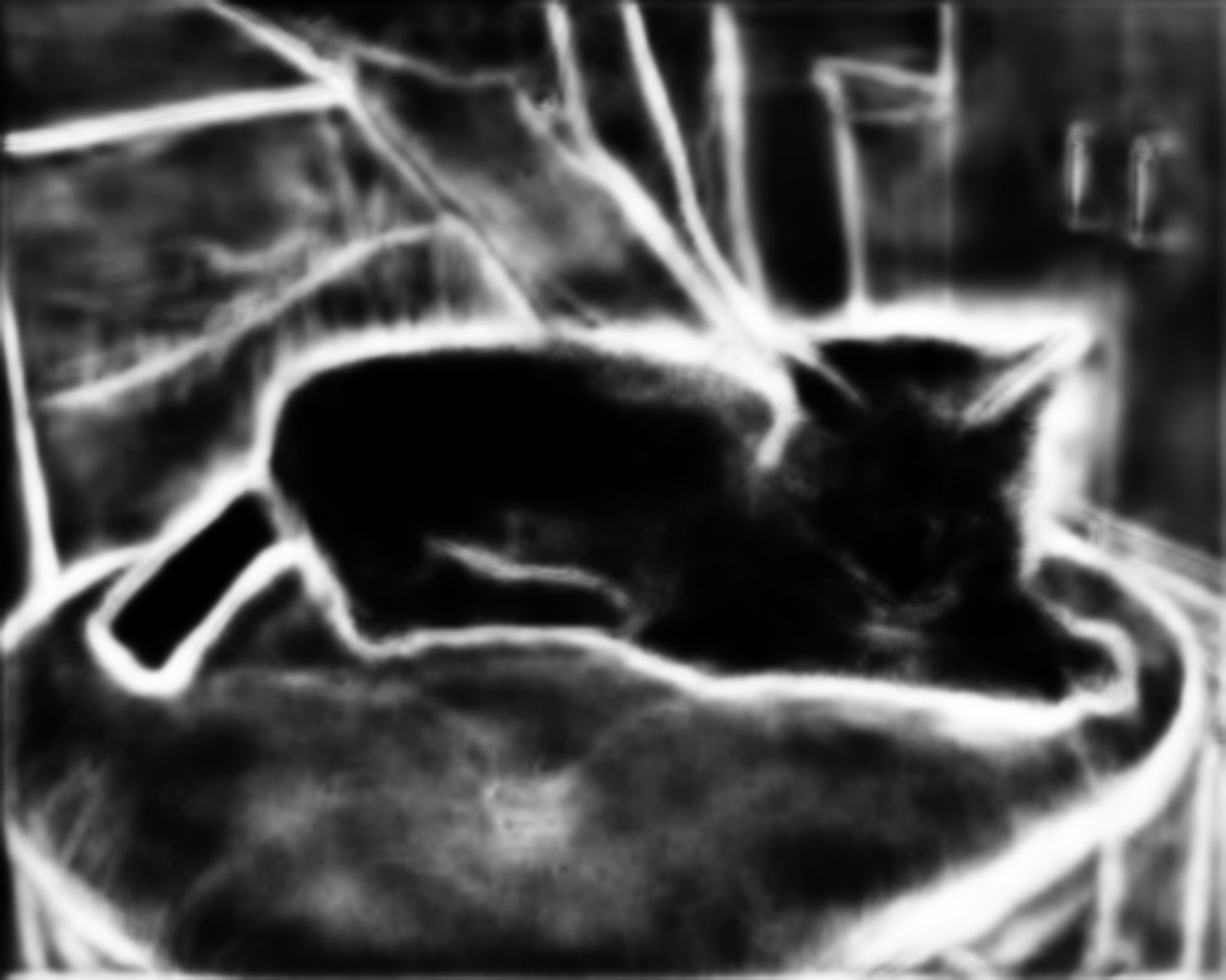}
\myfigurethreecol{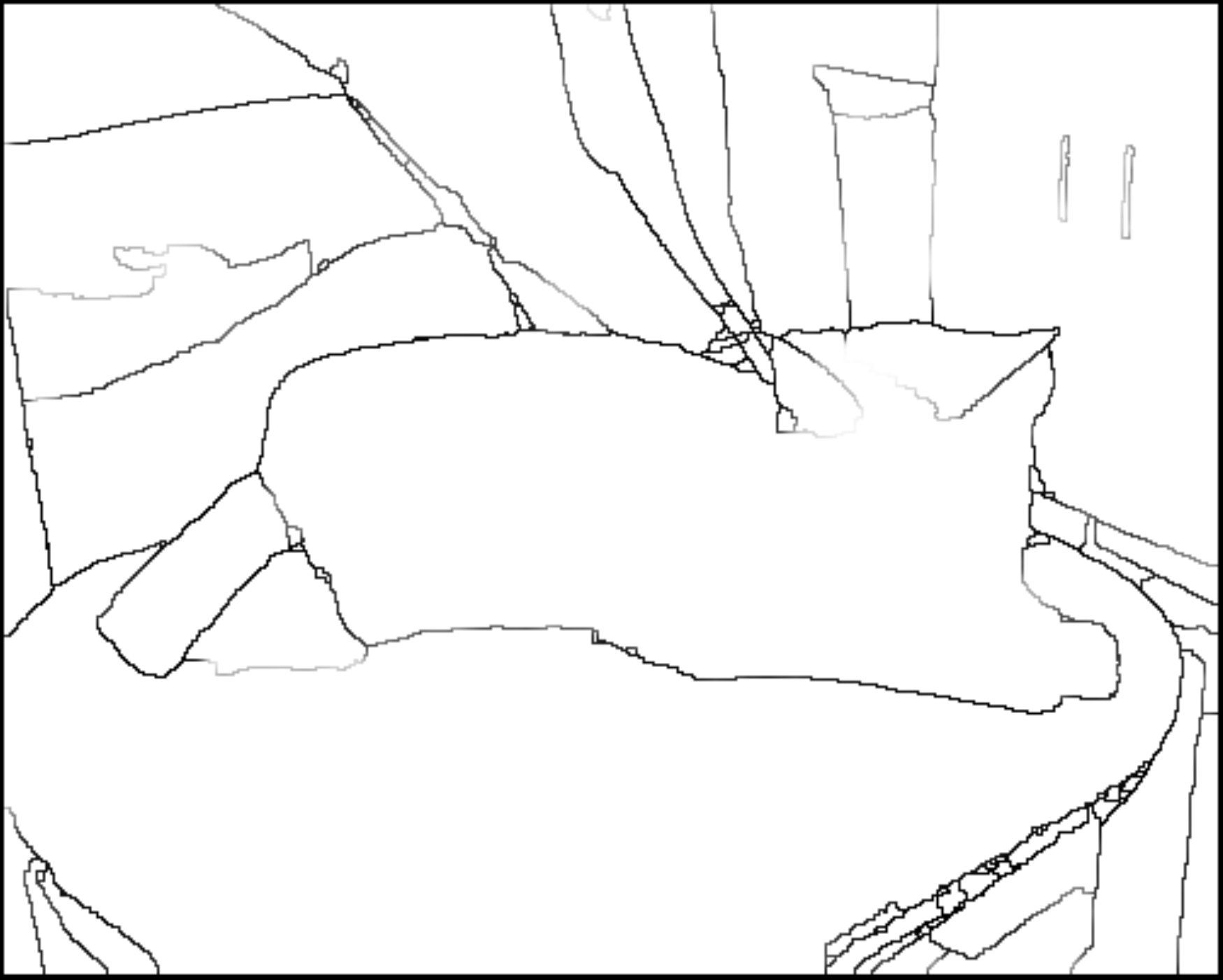}

\myfigurethreecol{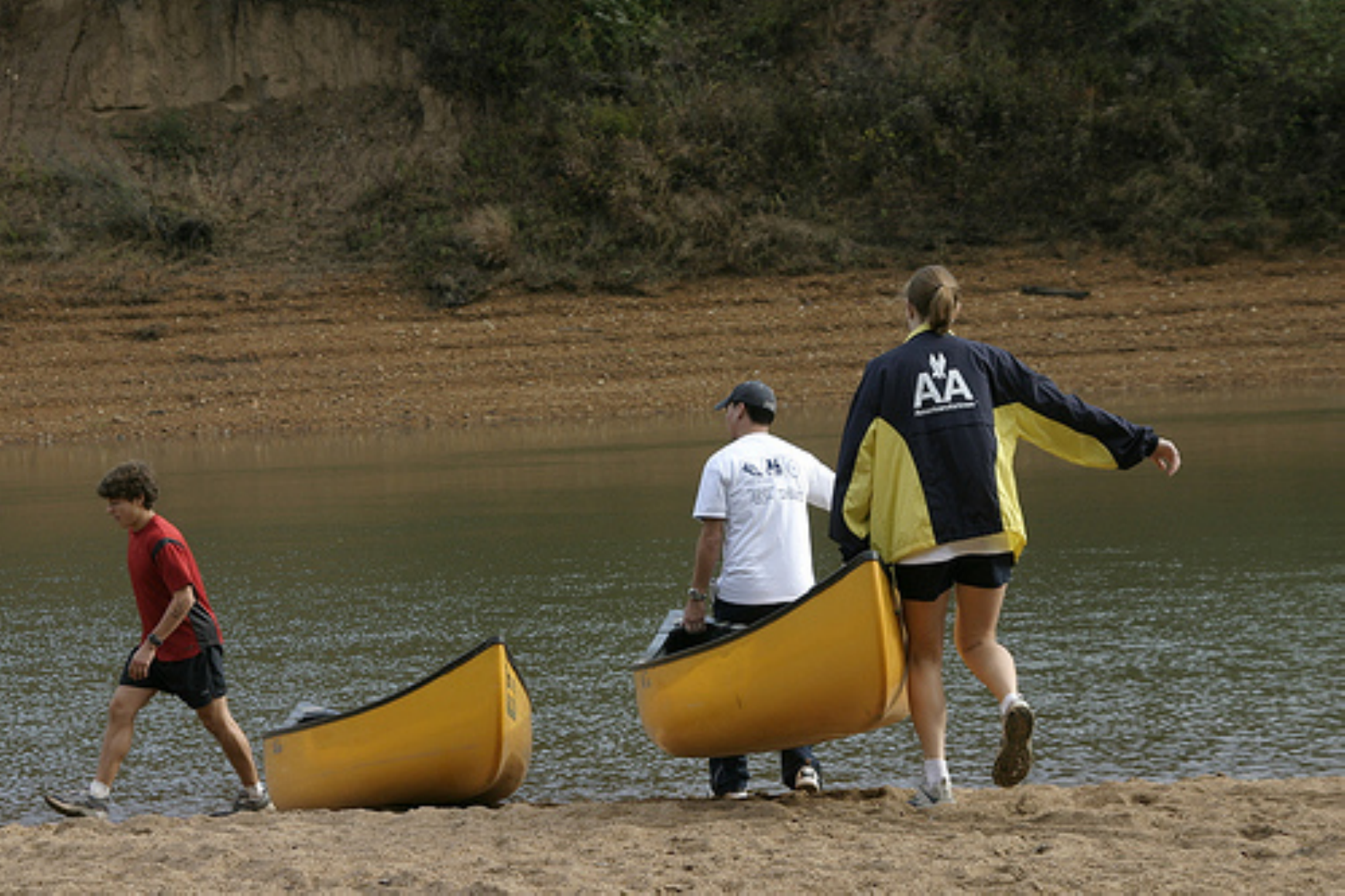}
\myfigurethreecol{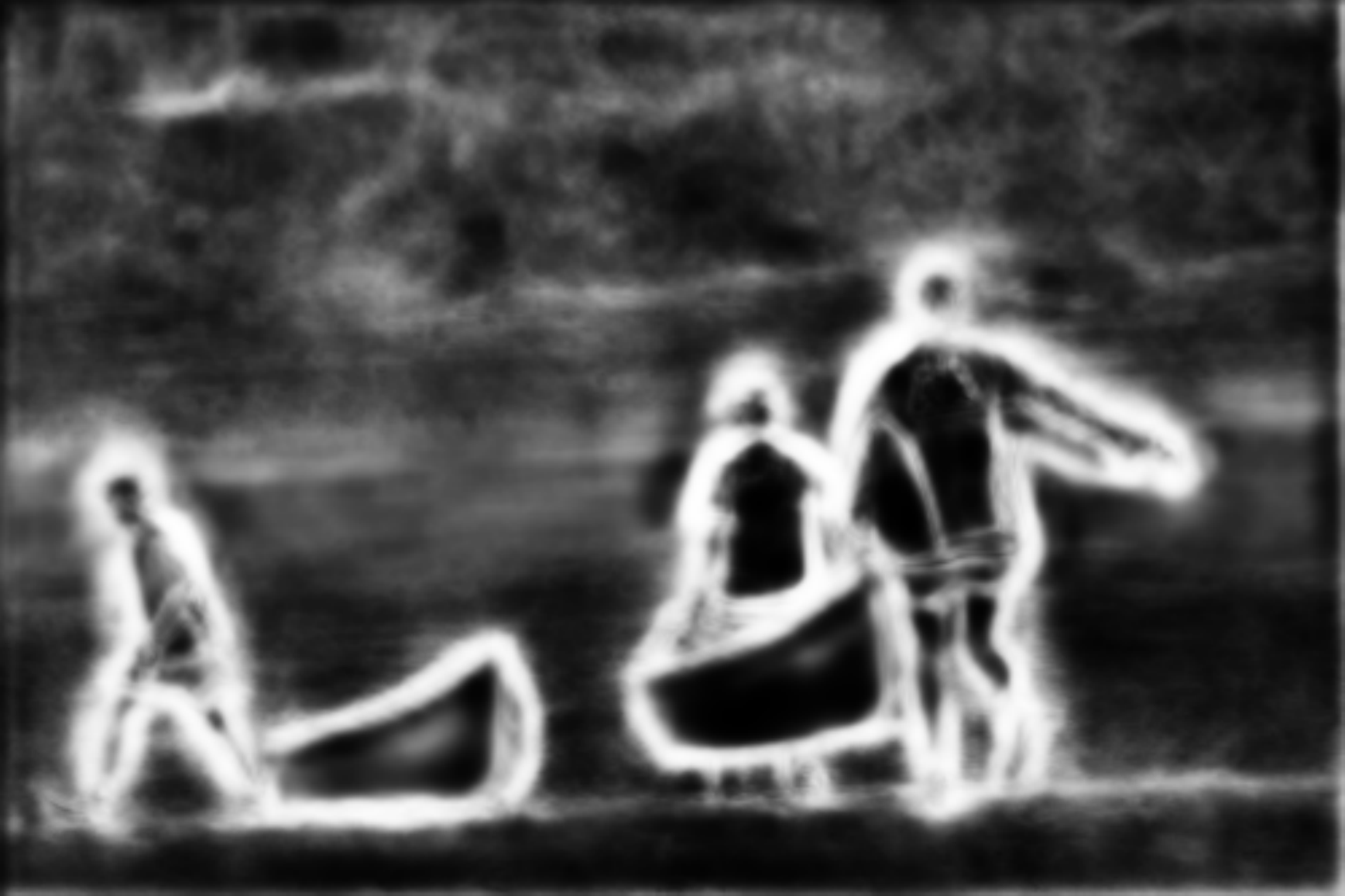}
\myfigurethreecol{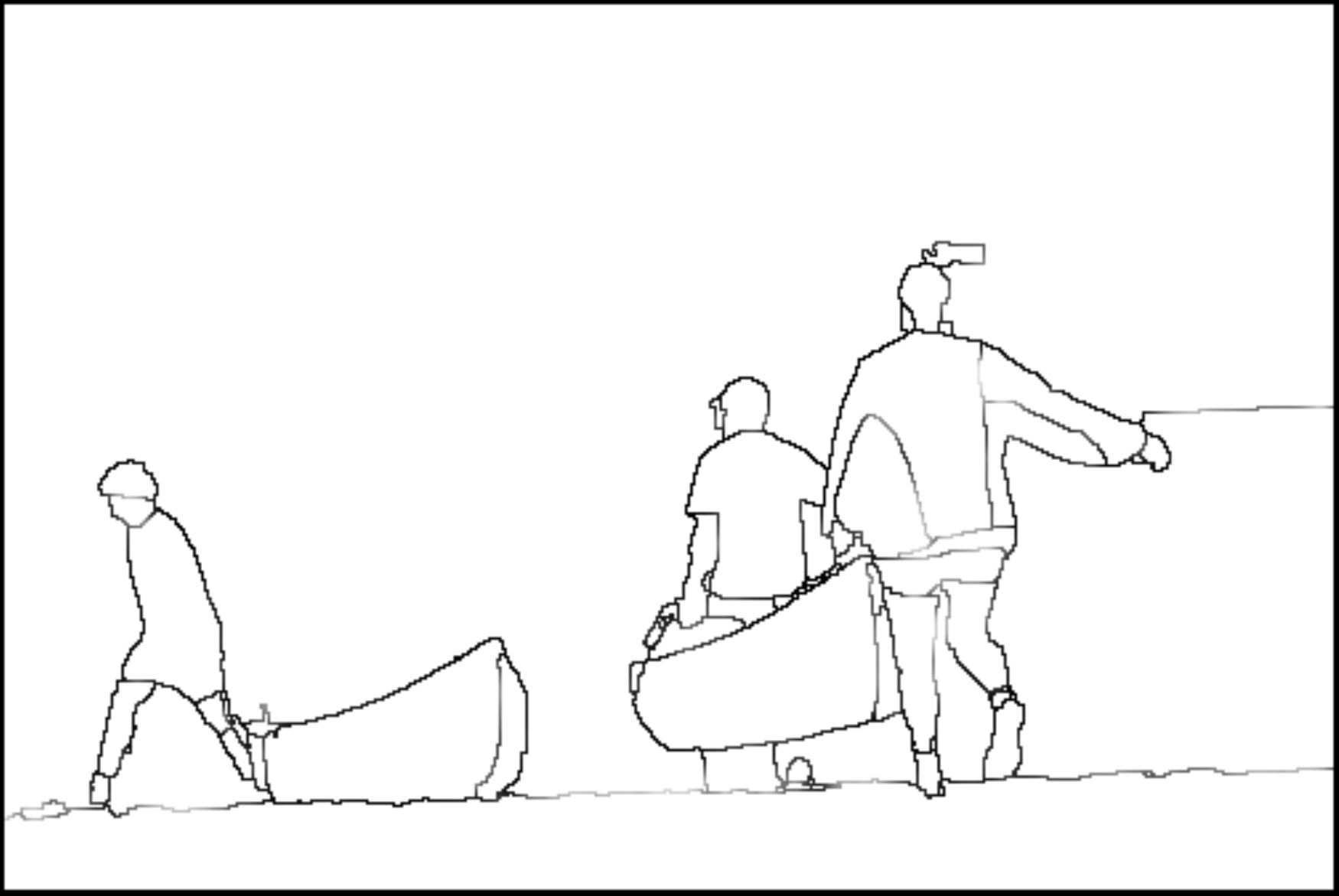}

\captionsetup{labelformat=default}
%\setcounter{figure}{1}
   %\caption{AA}
    \caption{A figure illustrating our boundary detection results. In the second column, we visualize the raw probability output of our boundary detector. In the third column, we present the final boundary maps after non-maximum suppression. While most prior methods predict the boundaries where the sharpest change in color occurs, our method captures semantic object-level boundaries, which we subsequently use to aid semantic segmentation.\vspace{-0.5cm}}
    \label{raw_con}
\end{figure}

\subsubsection{Constructing Pairwise Pixel Affinities.} We can use the predicted boundaries to build pairwise pixel affinities. Intuitively, we declare two pixels as similar (i.e., likely to belong to the same segment) if there is no boundary crossing the straight path between these two pixels. Conversely, two pixels are dissimilar if there is a boundary crossing their connecting path. The larger the boundary magnitude of the crossed path, the more dissimilar the two pixels should be, since a strong boundary is likely to mark the separation of two distinct segments. Similarly to~\cite{Arbelaez:2011:CDH:1963053.1963088}, we encode this intuition with a following formulation:

\begin{equation}
w^{sb}_{ij}=\exp{(\frac{-M_{ij}}{\sigma_{sb}})}
\end{equation} 

where $M_{ij}$ denotes the maximum boundary value that crosses the straight line path between pixels $i$ and $j$, $\sigma_{sb}$ depicts the smoothing parameter and $w^{sb}_{ij}$ denotes the semantic boundary-based affinity between pixels $i$ and $j$.

Similarly, we want to exploit high-level object information in the network to define another type of pixel similarity. Specifically, we use object class probabilities from the \textit{softmax} (SM) layer to achieve this goal. Intuitively, if pixels $i$ and $j$ have different hard segmentation labels from the \textit{softmax} layer, we set their similarity ( $w^{sm}_{ij}$) to $0$. Otherwise, we compute their similarity using the following equation:

\begin{equation} 
%w^{fc8}_{ij}=\exp{(\frac{-(fc8_{max}(i)-fc8_{max}(j))}{\sigma_{fc8}})}
w^{sm}_{ij}=\exp{(\frac{-D_{ij}}{\sigma_{sm}})}
\end{equation} 

where $D_{ij}$ denotes the difference in \textit{softmax} output values corresponding to the most likely object class for pixels $i$ and $j$,  and $\sigma_{sm}$ is a smoothing parameter. Then we can write the final affinity measure as:

\begin{equation} \label{eq:aff}
w_{ij}=\exp{(w^{sm}_{ij})}w^{sb}_{ij}
\end{equation}

We exponentiate the term corresponding to the object-level affinity because our boundary-based affinity may be too aggressive in declaring two pixels as dissimilar. To address this issue, we increase the importance of the object-level affinity in~\eqref{eq:aff} using the exponential function. However, in the experimental results section, we demonstrate that most of the benefit from modeling pairwise potentials comes from $w^{sb}_{ij}$ rather than $w^{sm}_{ij}$.

%Incorporating object-affinity term $w^{sm}_{ij}$ into Eq.~\eqref{eq:aff} provides only marginal improvements. 

We then use this pairwise pixel affinity measure to build a global affinity matrix  $W$ that encodes relationships between pixels in the entire image. For a given pixel, we sample $\approx 10\%$ of points in the neighborhood of radius $20$ around that pixel, and store the resulting affinities into $W$. 

%This results in a large and sparse affinity matrix. We note that even despite the sparsity of our affinity matrix, our proposed method still achieves solid results. 

%Additionally, we observe that constructing an affinity matrix is a pretty robust procedure and the results are not particularly sensitive to sampling fraction and neighborhood radius parameters. 

\subsection{Global Inference}

%explain that we decompose the SS problem into bunch of binary ones
% explain that closed form solution is continuous
%menton manifold ranking loss
% say that such formulation produces better results than commonly used globalization ntechniques

The last step in our proposed method is to combine semantic boundary information with the coarse segmentation from the FCN \textit{softmax} layer to produce an improved segmentation. We do this by introducing a global energy function that utilizes the affinity matrix constructed in the previous section along with the segmentation from the FCN \textit{softmax} layer. Using this energy, we perform a global inference to get segmentations that are well localized around the object boundaries and that are also spatially smooth.

Typical globalization models such as MRFs~\cite{Tappen:2003:CGC:946247.946707}, CRFs~\cite{NIPS2011_4296} or Graph Cuts~\cite{Boykov:2001:FAE:505471.505473}  produce a discrete label assignment for the segmentation problem by jointly modeling a multi-label distribution and solving a non-convex optimization. The common problem in doing so is that the optimization procedure may get stuck in local optima.

We introduce a new global energy function, which overcomes this issue and achieves better segmentation in comparison to prior globalization models. Similarly to prior globalization approaches, our goal is to minimize the energy corresponding to the sum of unary and pairwise potentials. However, the key difference in our approach comes from the relaxation of some of the constraints. Specifically, instead of modeling our problem as a joint multi-label distribution, we propose to decompose it into multiple binary problems, which can be solved concurrently. This decomposition can be viewed as assigning pixels to foreground and background labels for each of the different object classes. Additionally, we relax the integrality constraint. Both of these relaxations make our problem more manageable and allow us to formulate a global energy function that is differentiable, and has a closed form solution.

%In 33, the authors introduce the idea of global manifold ranking, and here we exploit in the context of semantic segmentation. Or present it and acknowledge at the end

In~\cite{NIPS2003_2506}, the authors introduce the idea of learning with global and local consistency in the context of semi-supervised problems. Inspired by this work, we incorporate some of these ideas in the context of semantic segmentation. Before defining our proposed global energy function, we introduce some relevant notation.

\captionsetup{labelformat=empty}
\captionsetup[figure]{skip=5pt}

\begin{figure*}
\centering

\myfigurefivecol{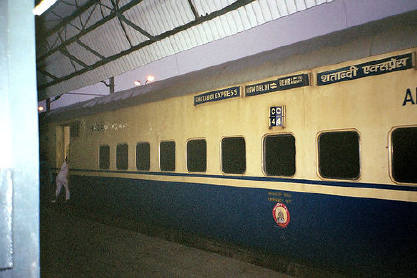}
\myfigurefivecol{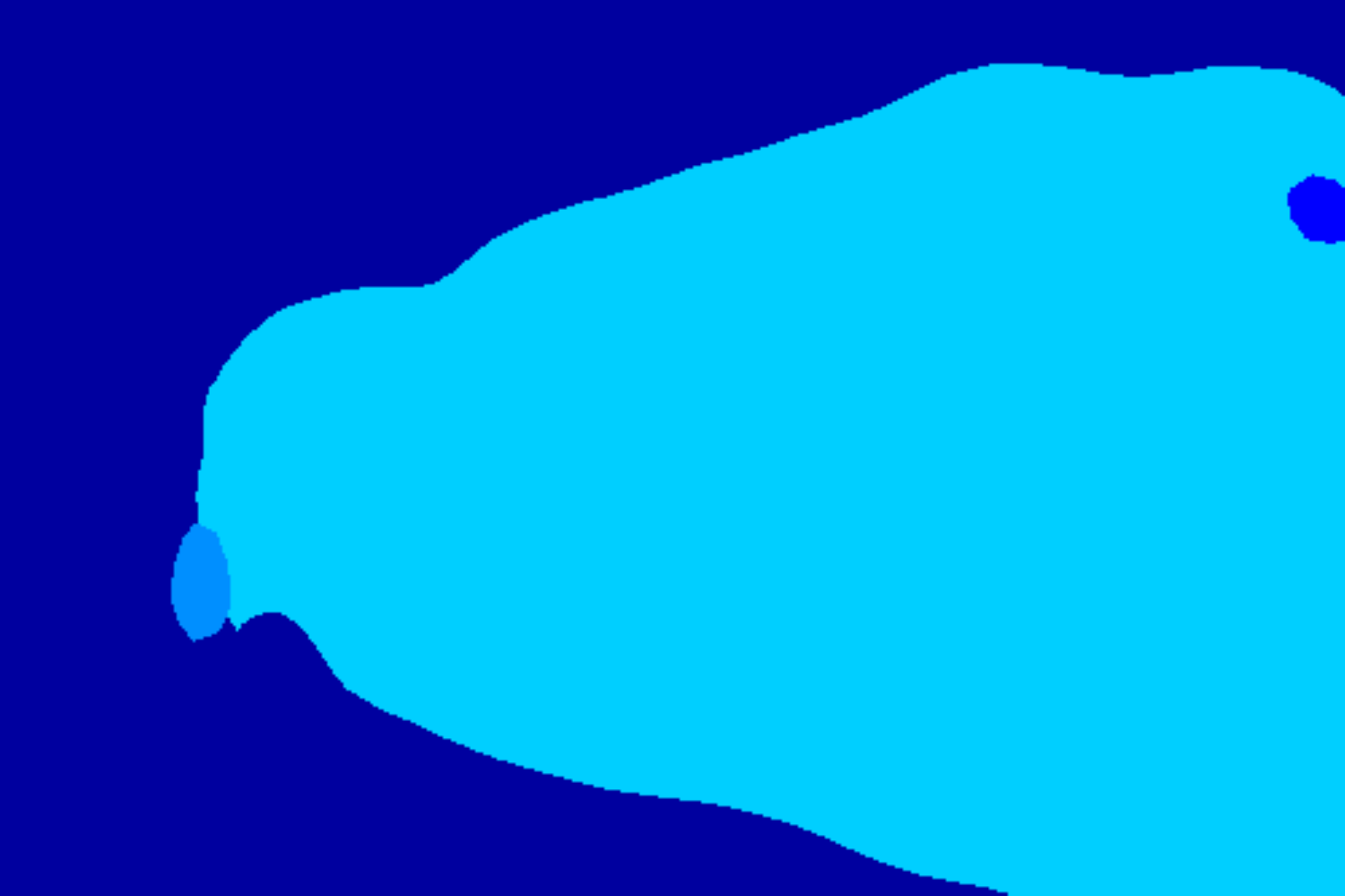}
\myfigurefivecol{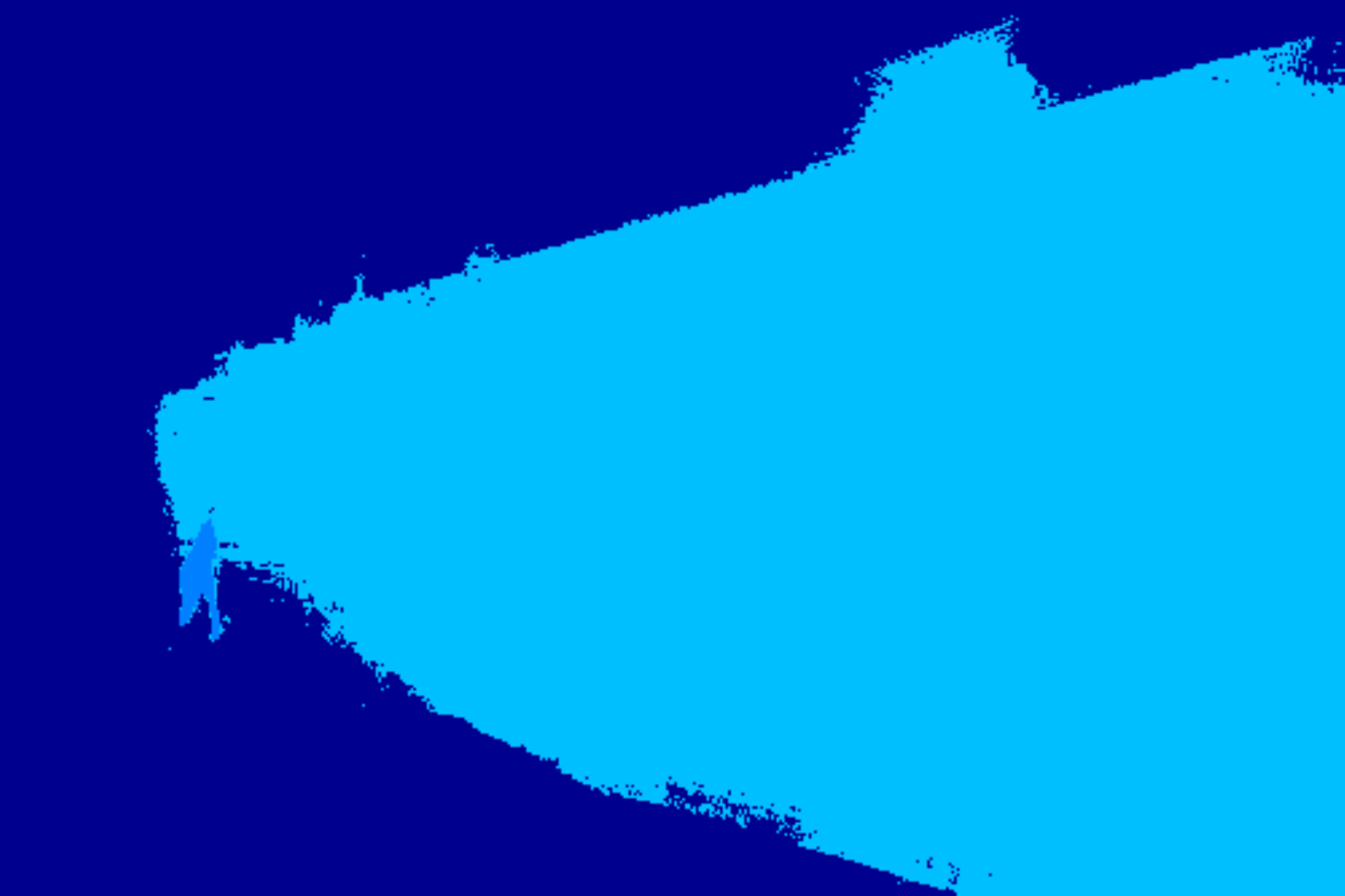}
\myfigurefivecol{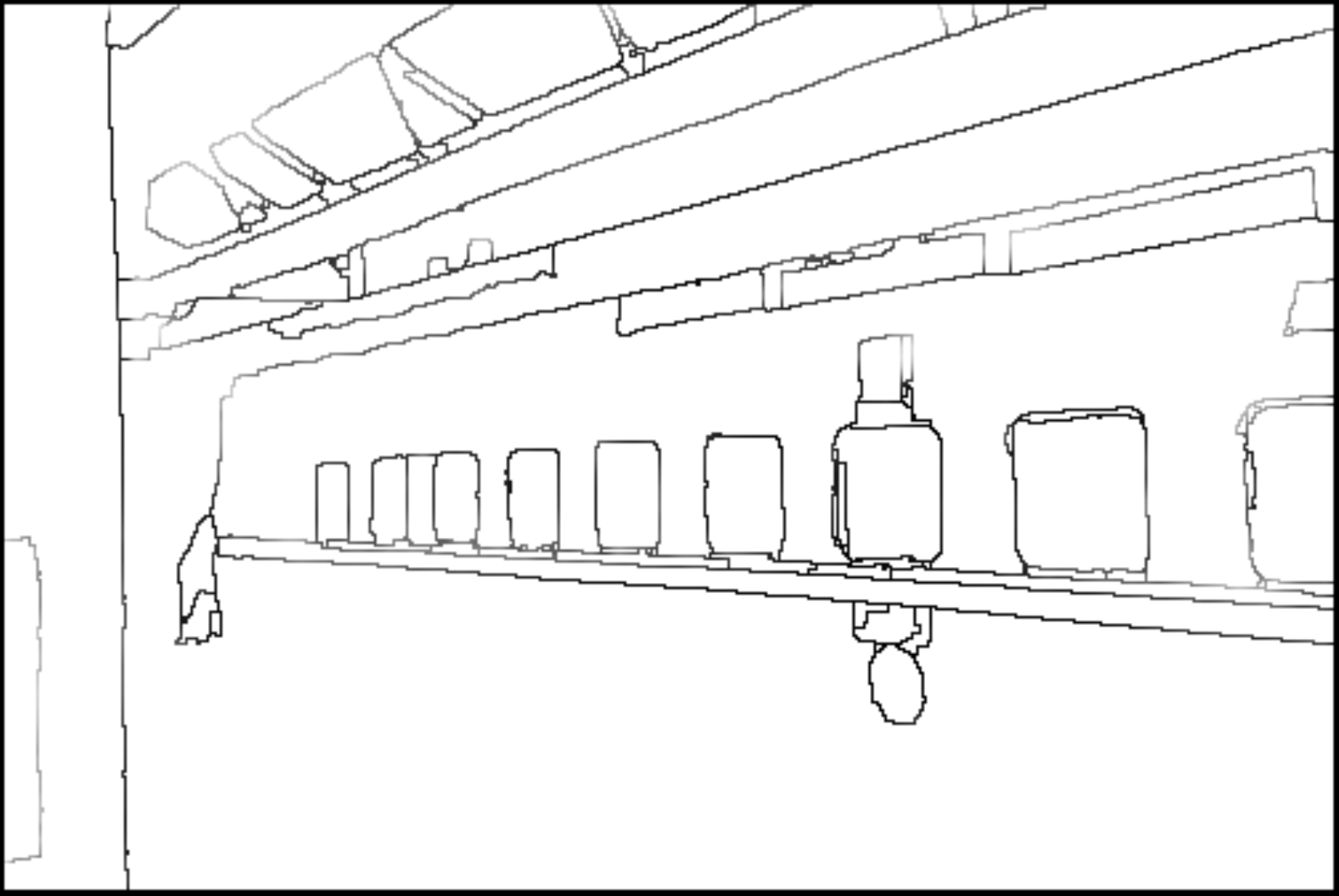}
\myfigurefivecol{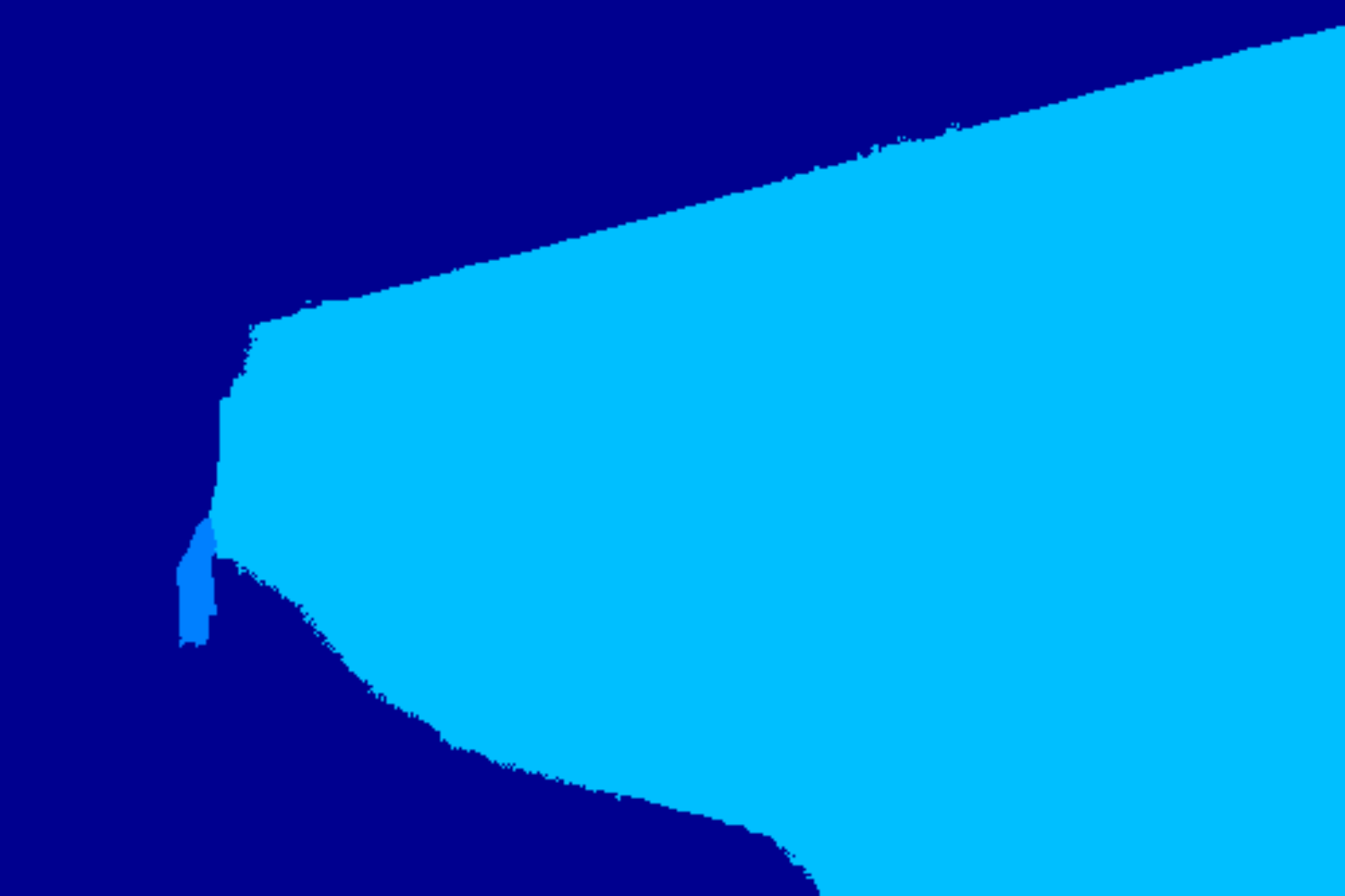}

\myfigurefivecolcaption{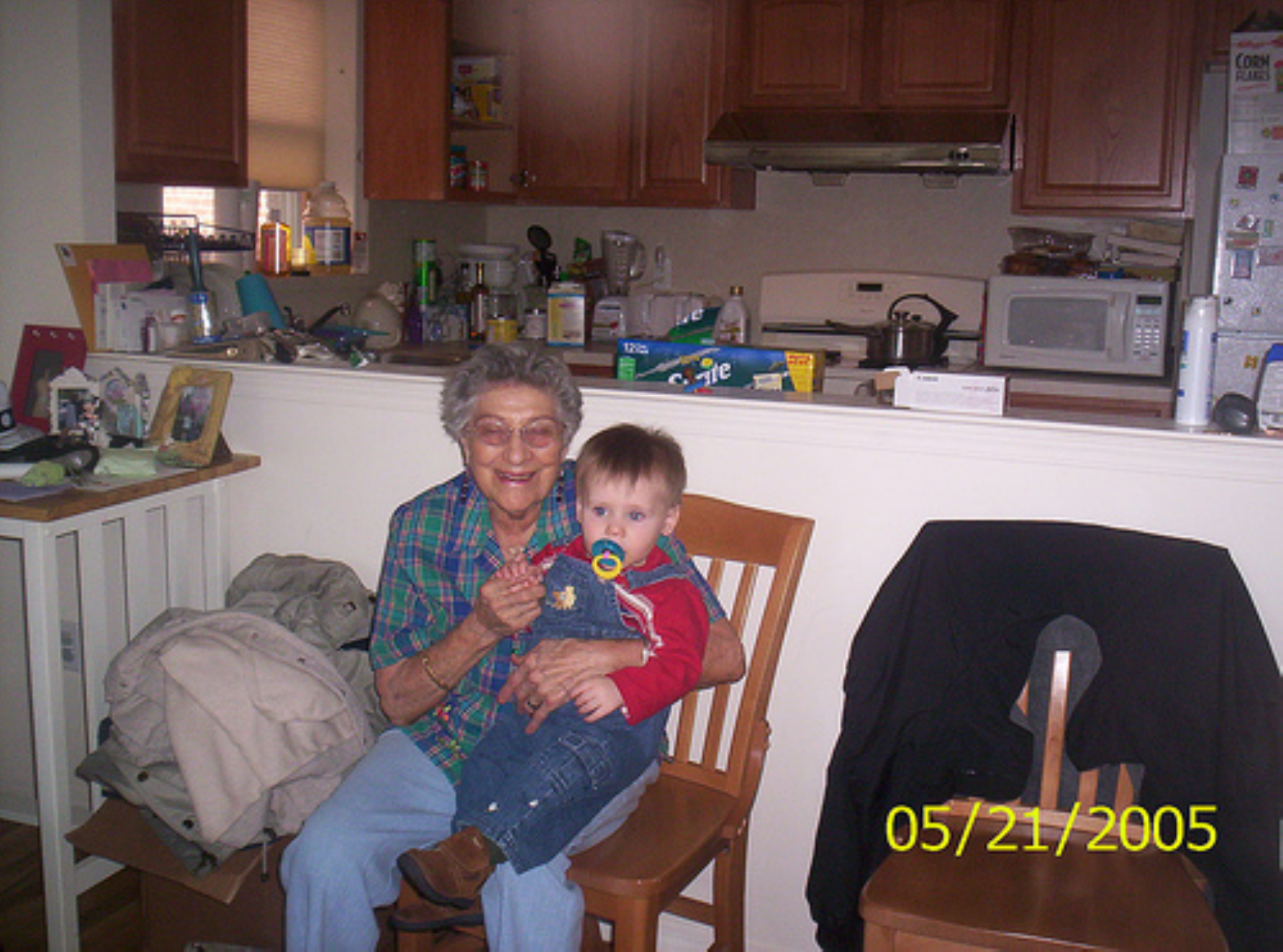}{Input}
\myfigurefivecolcaption{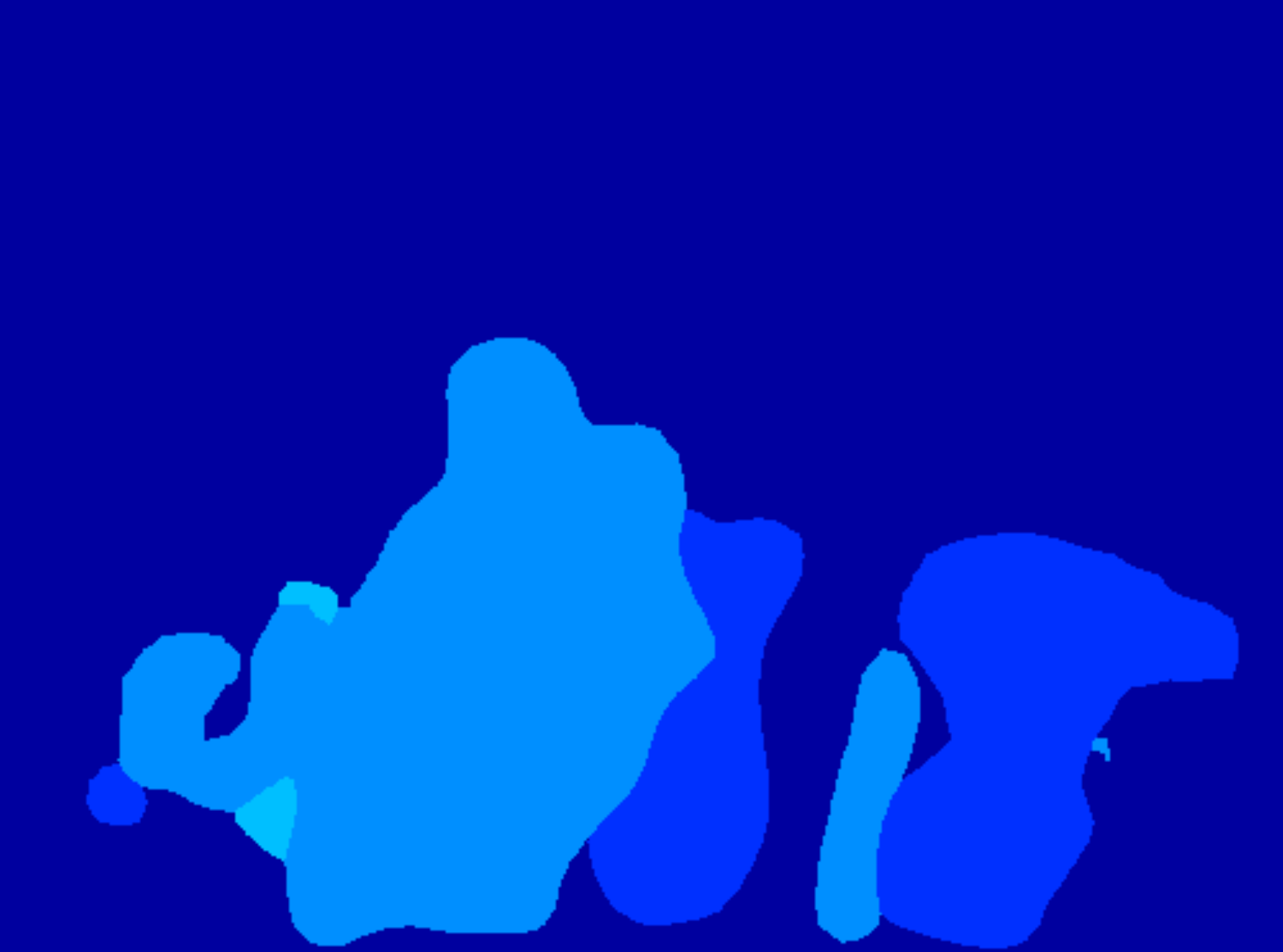}{Softmax}
\myfigurefivecolcaption{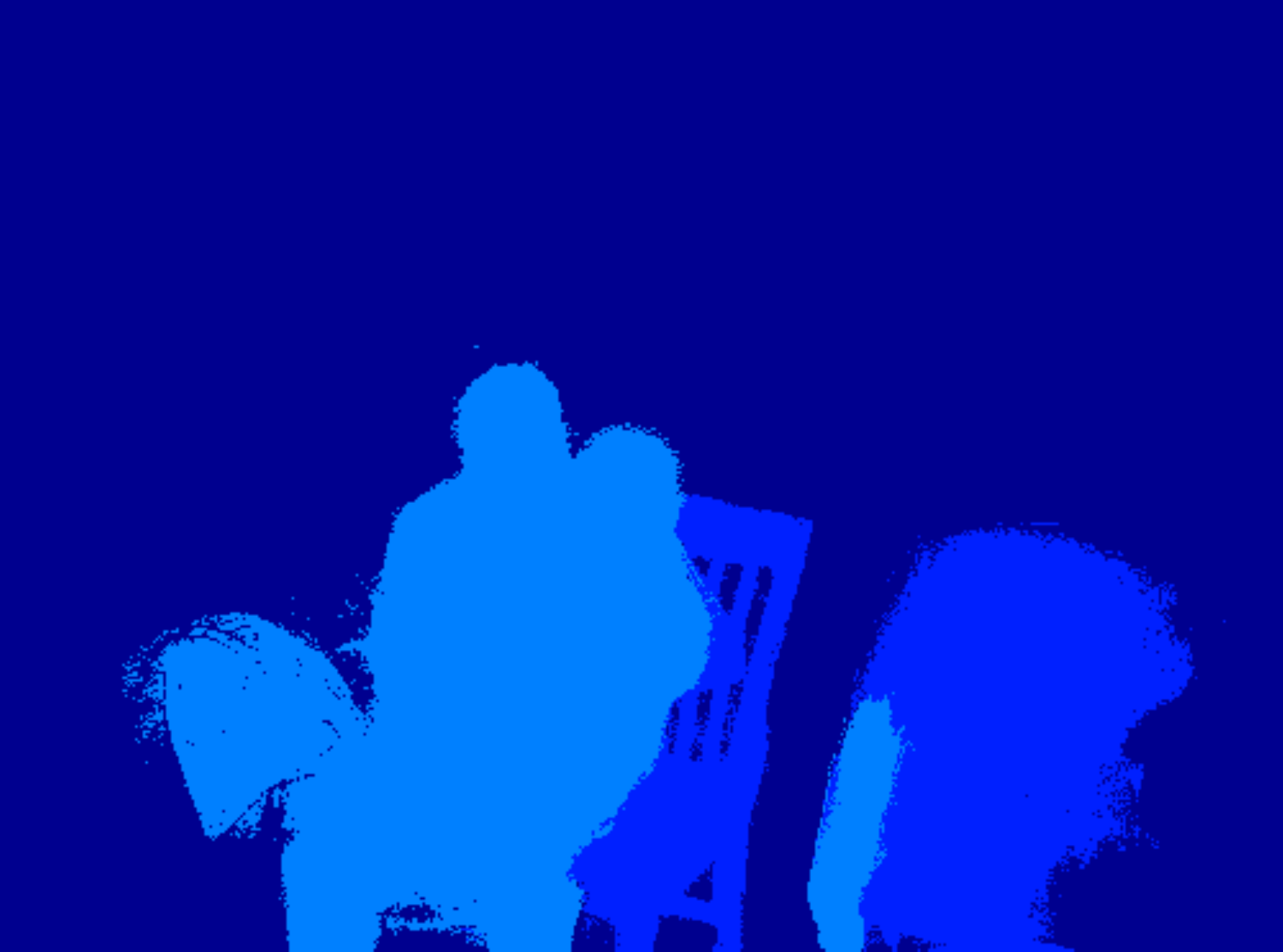}{Dense-CRF}
\myfigurefivecolcaption{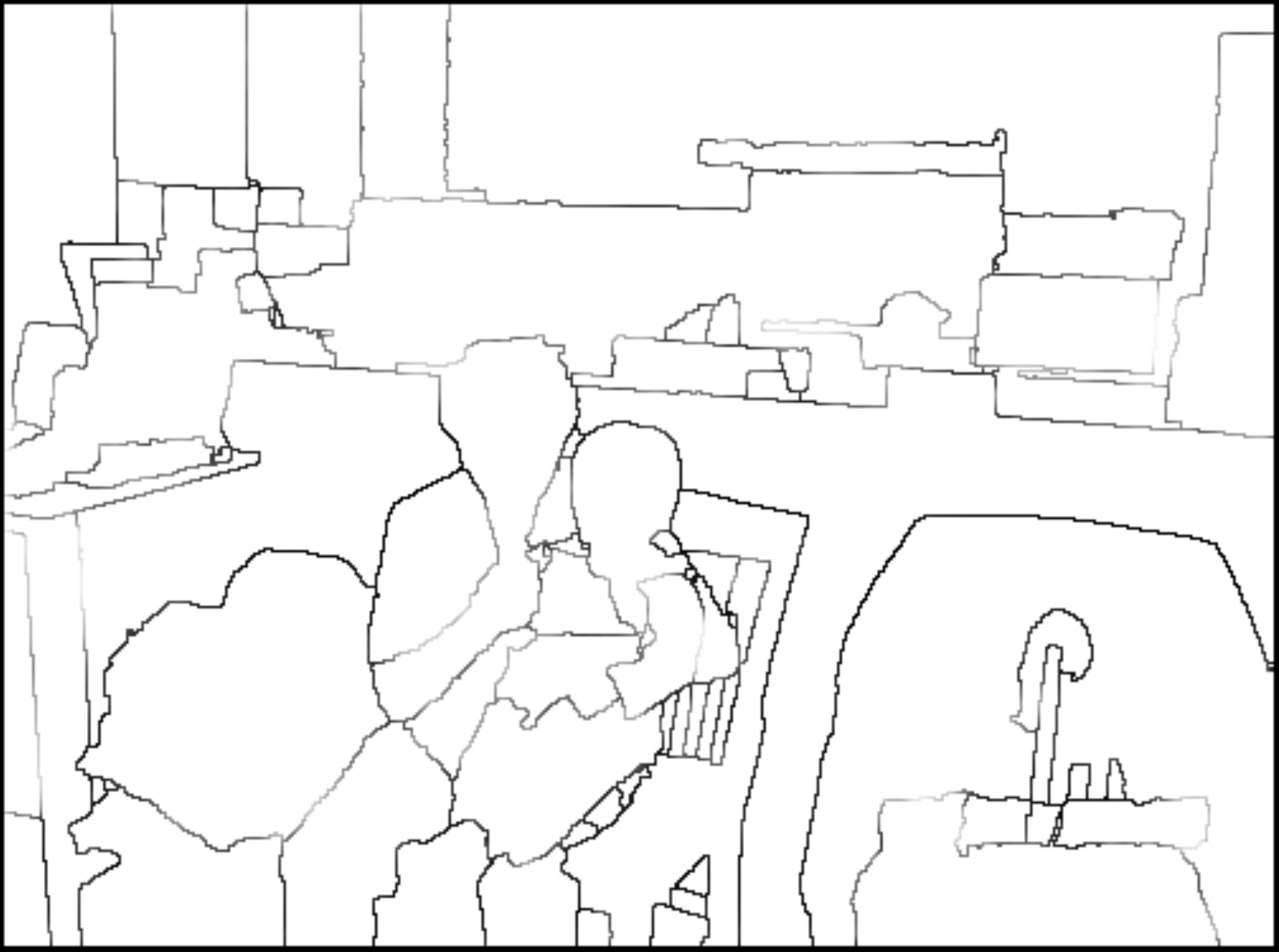}{BNF Boundaries}
\myfigurefivecolcaption{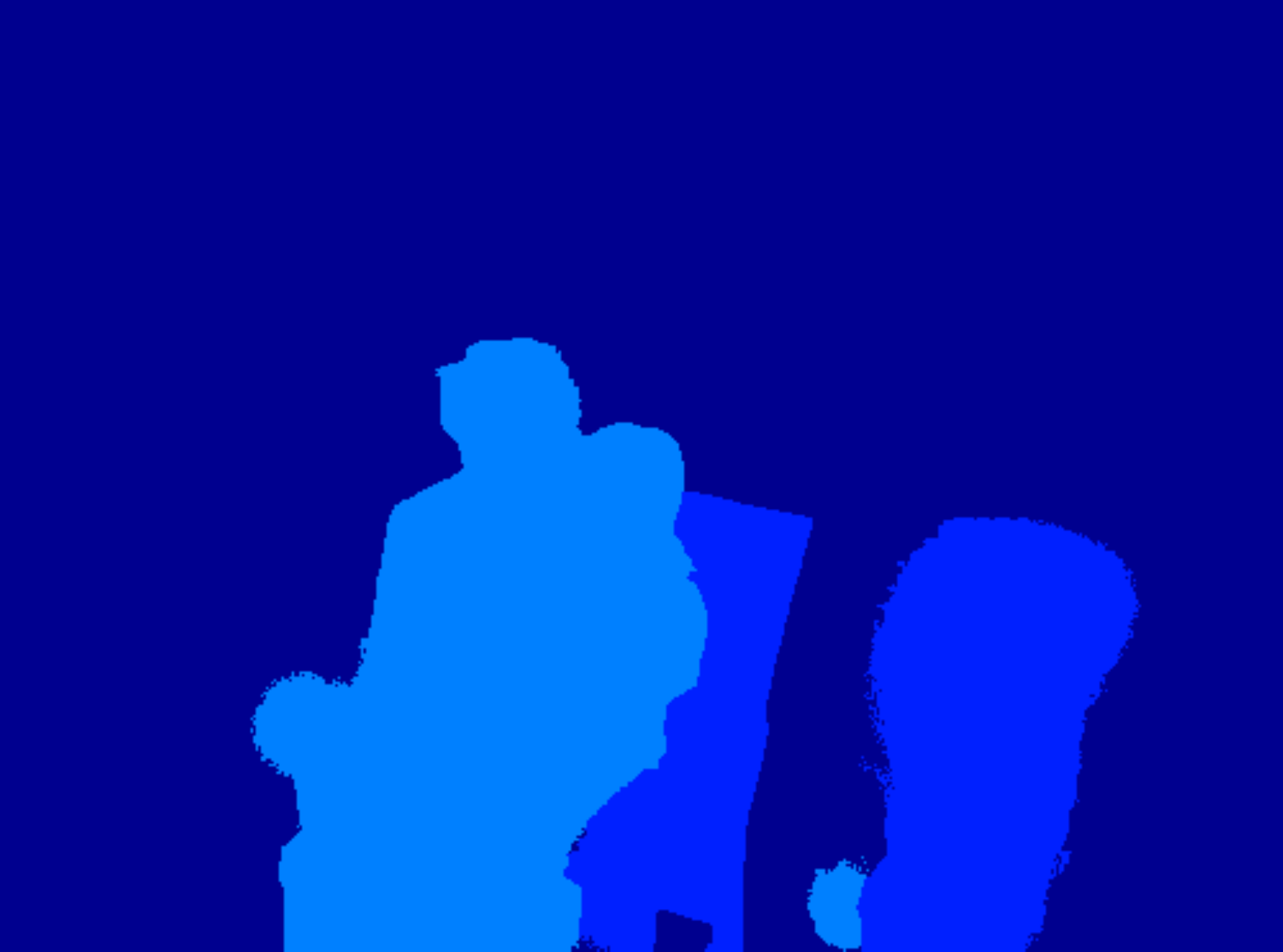}{BNF Segmentation}

\captionsetup{labelformat=default}
\setcounter{figure}{4}
   %\caption{AA}
    \caption{A figure illustrating semantic segmentation results. Images in columns two and three represent FCN~\textit{softmax} and Dense-CRF predictions, respectively. Note that all methods use the same FCN unary potentials. Additionally, observe that unlike FCN and Dense-CRF, our methods predicts segmentation that are both well localized around object boundaries and that are also spatially smooth.}
    \label{SS}
\end{figure*}

\captionsetup{labelformat=default}
\captionsetup[figure]{skip=10pt}

%%%% ENDED HERE %%%%%

%Additionally, we borrow some terminology from graph theory.  
For the purpose of illustration, suppose that we only have two classes: foreground and background. Then we can denote an optimal continuous solution to such a segmentation problem with variable $z^{*}$. To denote similarity between pixels $i$ and $j$ we use $ w_{ij}$. Then, $d_i$ indicates the degree of a pixel $i$. In graph theory, the degree of a node denotes the number of edges incident to that node. Thus, we set the degree of a pixel to $d_i=\sum_{j=1}^n w_{ij}$ for all $j$ except $i \neq j$. Finally, with $f_i$ we denote an initial segmentation probability, which in our case is obtained from the FCN \textit{softmax} layer.

%Finally, $\mu$ variable  balances the unary and pairwise terms in the equation (we set it to $0.025$).

Using this notation, we can then formulate our global inference objective as:

\begin{equation} \label{eq:loss}
z^{*}=\arg\!\min_z \frac{\mu}{2} \sum_i d_i (z_i -\frac{f_i}{d_i})^2 + \frac{1}{2} \sum_{ij} w_{ij}(z_i-z_j)^2
%z^{*}=\arg\!\min_z \frac{\mu}{2} \sum_i (d_iz_i -f_i)^2 + \frac{1}{2} \sum_{ij} w_{ij}(z_i-z_j)^2
\end{equation}

%The intuition here is that if the degree of a pixel $i$ is high, then assigning a new value $z_i$ which differs by a lot from $f_i$ should produce a large cost because it would also affect all the other similar pixels in the neighborhood of pixel $i$

This energy consists of two different terms. Similar to the general globalization framework, our first term encodes the unary energy while the second term includes the pairwise energy. We now explain the intuition behind each of these terms. The unary term attempts to find a segmentation assignment ($z_i$) that deviates little from the initial candidate segmentation computed from the \textit{softmax} layer (denoted by $f_i$). The $z_i$ in the unary term is weighted by the degree $d_i$ of the pixel in order to produce larger unary costs for pixels that have many similar pixels within the neighborhood. Instead, the pairwise term ensures that pixels that are similar should be assigned similar $z$ values. To balance the energies of the two terms we introduce a parameter $\mu$ and set it to $0.025$ throughout all our experiments.

%The pairwise term in the equation ensures that similar pixels are assigned similar $z$ values.

We can also express the same global energy function in matrix notation:

\begin{equation} \label{eq:mat_loss}
\bf{z^{*}}=\arg\!\min_{\bf{z}} \frac{\mu}{2} \bf{D}(\bf{z}-\bf{D^{-1}}\bf{f})^{T}(\bf{z}- \bf{D^{-1}}\bf{f}) + \frac{1}{2} \bf{z}^T(\bf{D}-\bf{W})\bf{z}
%\bf{z^{*}}=\arg\!\min_{\bf{z}} \frac{\mu}{2} (\bf{D}\bf{z}- \bf{f})^{T}(\bf{D}\bf{z}- \bf{f}) + \frac{1}{2} \bf{z}^T(\bf{D}-\bf{W})\bf{z}
%\arg\!\min_{\bf{Z}} \frac{\mu}{2} (\bf{D}\bf{Z}- \bf{F})^{T}(\bf{D}\bf{Z}- \bf{F}) + \frac{1}{2} \bf{Z}^T(\bf{D}-\bf{W})\bf{Z}
%\frac{\mu}{2} (\bf{D}\bf{Z}- \bf{F})^{T}(\bf{D}\bf{Z}- \bf{F}) + \frac{1}{2} \bf{Z}^T(\bf{D}-\bf{W})\bf{Z}
\end{equation} 

where $\bf{z^{*}}$ is a $n \times 1$ vector containing an optimal continuous assignment for all $n$ pixels, $\bf{D}$ is a diagonal degree matrix, and $\bf{W}$ is the $n \times n$ pixel affinity matrix. Finally, $\bf{f}$ denotes a $n \times 1$ vector containing the probabilities from the \textit{softmax} layer corresponding to a particular object class. 

An advantage of our energy is that it is differentiable. If we denote the above energy as $E(z)$ then the derivative of this energy can be written as follows:

\begin{equation} \label{eq:deriv}
%\frac{\partial{E(z)}}{\partial{z}}= \mu(\bf{D}\bf{z}- \bf{f}) + (\bf{D}-\bf{W})\bf{z} =0
\frac{\partial{E(z)}}{\partial{z}}= \mu\bf{D}(\bf{z}- \bf{D^{-1}}\bf{f}) + (\bf{D}-\bf{W})\bf{z} =0
\end{equation} 

With simple algebraic manipulations we can then obtain a closed form solution to this optimization:

\begin{equation} \label{eq:sol}
\bf{z^{*}}=(\bf{D}-\alpha \bf{W})^{-1} \beta \bf{f}
\end{equation} 

where $\alpha = \frac{1}{1+\mu}$ and $\beta = \frac{\mu}{1+\mu}$. In the general case where we have $k$ object classes we can write the solution as:

\begin{equation} \label{eq:mat_sol}
\bf{Z^{*}}=(\bf{D}-\alpha \bf{W})^{-1} \beta \bf{F}
\end{equation} 

where $\bf{Z}$ now depicts a $n \times k$ matrix containing assignments for all $k$ object classes, while $\bf{F}$  denotes $n \times k$ matrix with object class probabilities from \textit{softmax} layer. Due to the large size of $\bf{D}-\alpha \bf{W}$ it is impractical to invert it. However,  if we consider an image as a graph where each pixel denotes a vertex in the graph, we can observe that the term $\bf{D}-\bf{W}$ in our optimization is equivalent to a Laplacian matrix of such graph. Since we know that a Laplacian matrix is positive semi-definite, we can use the preconditioned conjugate gradient method~\cite{Shewchuk:1994:ICG:865018} to solve the system in Eq.~\eqref{eq:axb}. Alternatively, because our defined global energy in Eq.~\eqref{eq:mat_loss} is differentiable, we can efficiently solve this optimization problem using stochastic gradient descent. We choose the former option and solve the following system:

\begin{equation} \label{eq:axb}
(\bf{D}-\alpha \bf{W})\bf{z^{*}}=\beta \bf{f}
\end{equation} 

To obtain the final discrete segmentation, for each pixel we assign the object class that corresponds to the largest column value in the row of $\bf{Z}$ (note that each row in $\bf{Z}$ represents a single pixel in the image, and each column in $\bf{Z}$ represents one of the object classes). In the experimental section, we show that this solution produces better quantitative and qualitative results in comparison to commonly used globalization techniques. 

%Additionally, due to the differentiability of this energy we could also train the network in the front-to-end fashion. The fact that we outperform commonly used globalization methods using just the inference step, suggests that we could achieve even more substantial accuracy gains by optimizing the full network with respect to this energy. We will investigate this possibility in our future work.

%It can be seen as a $n \times k$ matrix where $n$ denotes the number of pixels in the image and $k$ depicts the number of object classes.

\section{Experimental Results}

  \setlength{\tabcolsep}{2pt}

     \begin{table}
     %\footnotesize
     \scriptsize
    \begin{center}
    %\begin{tabular}{ | c | C{0.15cm} | C{0.15cm} | C{0.15cm} | C{0.15cm} | C{0.15cm} | C{0.15cm} | C{0.15cm} | C{0.15cm} | C{0.15cm} | C{0.15cm} | C{0.15cm} | C{0.15cm} | C{0.15cm} | C{0.15cm} | C{0.15cm} | C{0.15cm} | C{0.15cm} | C{0.15cm} | C{0.15cm} | C{0.15cm} ? C{0.15cm} |}
    \begin{tabular}{ | c | c | c | c |}
    %\cline{2-9}
    \hline
    %Method (Metric) & Inverse Detectors (MF) & Inverse Detectors-HOG (MF) & ObjEdge (MF) \\ \hline\hline
     Metric & Inference Method & RGB Affinity & \bf  BNF Affinity \\ \hline
        \multirow{5}{*}{PP-IOU}
	& Belief Propagation~\cite{Tappen:2003:CGC:946247.946707} & 75.4 & \bf 75.6\\ 
	& ICM & 74.2 & \bf 75.8\\
	& TRWS~\cite{TRWS} & 75.9 & \bf 76.7\\ 
	%& Dense-CRF~\cite{NIPS2011_4296} & \bf 77.3 & -\\	
	& QPBO~\cite{QPBO} & 76.9 & \bf 77.2\\ 
	%& Dense-CRF~\cite{NIPS2011_4296} &  77.3 & -\\	
	& \bf BNF & 74.6 & \bf 77.6\\  \Xhline{4\arrayrulewidth}
	 \multirow{5}{*}{PI-IOU}
	& Belief Propagation~\cite{Tappen:2003:CGC:946247.946707}  & 45.9 & \bf 46.2\\
	&  ICM  & 45.7 & \bf 48.8 \\
	& TRWS~\cite{TRWS} & 51.5 & \bf 52.0\\ 
	%&  Dense-CRF~\cite{NIPS2011_4296}  & \bf 51.1 & -\\
	& QPBO~\cite{QPBO} & 55.3 & \bf 57.2\\
	%&  Dense-CRF~\cite{NIPS2011_4296}  & 51.1 & -\\
	&  \bf BNF & 53.0 & \bf 58.5\\ \hline

      \end{tabular}
    \end{center}
    \caption{We compare semantic segmentation results when using a color-based pixel affinity and our proposed boundary-based affinity. We note that our proposed affinity improves the performance of all globalization techniques. Note that all of the inference methods use the \textbf{same FCN unary potentials}. This suggests that for every method our boundary-based affinity is more beneficial for semantic segmentation than the color-based affinity.}
     %\caption{Semantic segmentation results on the SBD and VOC 2007 datasets according to the PI-IOU evaluation metric. We denote the DeepLab-CRF system and our proposed modification as DL-CRF and DL-CRF+\HfL, respectively. According to the PI-IOU metric, our proposed features improve the mean accuracy by 3\% and $1.9\%$ on SBD and VOC 2007 datasets respectively.}
    \label{affinity_table}
   \end{table}

In this section we present quantitative and qualitative results for semantic segmentation on the SBD~\cite{BharathICCV2011} dataset, which contains objects and their per-pixel annotations for $20$ Pascal VOC classes. We evaluate semantic segmentation results using two evaluation metrics. The first metric measures accuracy based on pixel intersection-over-union averaged per pixels (PP-IOU) across the 20 classes. According to this metric, the accuracy is computed on a per-pixel basis. As a result, the images that contain large object regions are given more importance. However, for certain applications we may need to accurately segment small objects. Therefore, similar to~\cite{gberta_2015_ICCV} we also consider the PI-IOU metric (pixel intersection-over-union averaged per image across the 20 classes), which gives equal weight to each of the images.

We compare Boundary Neural Fields with other commonly used global inference methods. These methods include Belief Propagation~\cite{Tappen:2003:CGC:946247.946707}, Iterated Conditional Mode (ICM), Graph Cuts~\cite{Boykov:2001:FAE:505471.505473}, and Dense-CRF~\cite{NIPS2011_4296}. Note that in all of our evaluations we use the same FCN unary potentials for every model.

%Additionally, we also compare all of these methods with the output from FCN's \textit{softmax} layer. 

Our evaluations provide evidence for three conclusions:
\begin{itemize}
\item In Subsection~\ref{aff_results}, we show  that our boundary-based pixel affinities are better suited for semantic segmentation than the traditional color-based affinities.
\item In Subsection~\ref{inf_results}, we demonstrate that our global minimization leads to better results than those achieved by other inference schemes.
\item In Fig.~\ref{SS}, we qualitatively compare the outputs of FCN and Dense-CRF to our predicted segmentations. This comparison shows that the BNF segments are better localized around the object boundaries and that they are also spatially smooth.
\end{itemize}

%that using our proposed affinity leads to better segmentation results in comparison to the color based affinity in all the globalization techniques included in our comparisons. Additionally, that minimizing our proposed global energy produces better results than the results of other globalization techniques. Finally, we show .(See 

%This ensures that predictions for the smaller object regions become as important as for the large

 \captionsetup{labelformat=default}

  \setlength{\tabcolsep}{2pt}

     \begin{table*}
     %\footnotesize
     \scriptsize
    \begin{center}
    %\begin{tabular}{ | c | C{0.15cm} | C{0.15cm} | C{0.15cm} | C{0.15cm} | C{0.15cm} | C{0.15cm} | C{0.15cm} | C{0.15cm} | C{0.15cm} | C{0.15cm} | C{0.15cm} | C{0.15cm} | C{0.15cm} | C{0.15cm} | C{0.15cm} | C{0.15cm} | C{0.15cm} | C{0.15cm} | C{0.15cm} | C{0.15cm} ? C{0.15cm} |}
    \begin{tabular}{ | c | c | c | c | c | c | c | c | c | c | c | c | c | c | c | c | c | c | c | c | c | c ? c |}
    %\cline{2-9}
    \hline
    %Method (Metric) & Inverse Detectors (MF) & Inverse Detectors-HOG (MF) & ObjEdge (MF) \\ \hline\hline
     Metric & Method & aero & bike & bird & boat & bottle & bus & car & cat & chair & cow & table & dog & horse & mbike & person & plant & sheep & sofa & train & tv & mean\\ \hline
        \multirow{9}{*}{PP-IOU}
        	%& DL-CRF (VOC)& \bf 78.6 & 41.1 & \bf 83.5 & \bf 75.3 & 72.9 & 83.1 & \bf 76.6 & \bf 80.8 & \bf 37.8 & \bf 72.1 & 66.5 & \bf 64.7 & 65.8 & \bf 75.7 & \bf 80.5 & \bf 34.4 & \bf 75.9 & \bf 47.4 & 86.6 & \bf 77.9 & \bf 68.9\\ 	
	%& \bf DL-CRF+\HfL (VOC) & 77.9 & \bf 41.2 & 83.1 & 74.4 & \bf 73.2 & \bf 85.5 & 76.1 & 80.6 & 35.7 & 71.0 & \bf 66.6 & 64.3 & \bf 65.9 & 75.2 & 80.2 & 32.8 & 75.2 & 47.0 & \bf 87.1 & \bf 77.9 & 68.5\\  \cline{2-23}
	& FCN-Softmax & 80.7 & 71.6 & 80.7 & 71.3 & 72.9 & 88.1 & 81.8 & 86.6 & 47.4 & 82.9 & 57.9 & 83.9 & 79.6 & 80.4 & 81.0 & 64.7 & 78.2 & 54.5 & 80.9 & 69.9 &  74.8\\ 
	& Belief Propagation~\cite{Tappen:2003:CGC:946247.946707} & 81.4 & 72.2 & 82.4 & 72.2 & 74.3 & 88.8 & 82.4 & 87.2 & 48.4 & 83.8 & 58.4 & 84.6 & 80.5 & 80.9 & 81.5 & 65.1 & 79.5 & 55.5 &  81.5 & 71.2 & 75.6\\ 
	& ICM & 81.7 & 72.2 & 82.8 & 72.1 & 75.3 & 89.6 & 83.4 & 87.7 & 46.3 & 83.3 & 58.4 & 84.6 & 80.6 & 81.4 & 81.5 & 65.8 & 79.5 & 56.0 & 80.7 & 74.1 & 75.8\\
	& TRWS~\cite{TRWS} & 81.6 & 70.9 & 83.8 & 72.0 & 75.1 & 89.5 & 82.5 & 88.0 & \bf 51.7 & 86.6 & 61.9 & 85.8 & 83.3 & 80.8 & 81.1 & 65.3 & 81.5 & \bf 58.8 & 77.6 & 75.9 &  76.7\\
	& Graph Cuts~\cite{Boykov:2001:FAE:505471.505473} & 82.5 & 72.4 & 84.6 & 73.3 & \bf 77.2 & 89.7 & 83.3 & 88.8 & 49.3 & 84.0 & 60.3 & 85.4 & 82.2 & 81.2 & 81.9 & 66.7 & 79.8 & 58.0 & 82.3 & 74.9 &  76.9\\
	& QPBO~\cite{QPBO} & 82.6 & 72.3 & 84.7 & 73.1 & 76.7 & 89.9 & \bf 83.6 & 89.3 & 49.7 & 85.0 & 61.1 & 86.2 & 82.9 & 81.3 &\bf 82.3 & 67.1 & 80.5 & \bf 58.8 & 82.2 & 75.1 &  77.2\\
	& Dense-CRF~\cite{NIPS2011_4296} & \bf 83.4 & 71.5 & 84.9 & 72.6 & 76.2 & 89.5 & 83.3 & 89.1 & 50.4 & \bf 86.7 & 61.0 &\bf 86.8 & \bf 83.5 & \bf 81.8 & \bf 82.3 & 66.9 & \bf 82.2 & 58.2 & 81.9 & 75.1 &  77.3\\
	& \bf BNF-SB & 81.9 & 72.5 & 84.9 & 73.3 & 76.0 & 90.3 & 83.1 & 89.2 & 51.2 & \bf 86.7 & \bf 61.5 & 86.6 & 83.2 & 81.3 & 81.9 & 66.2 & 81.7 &  58.6 & 81.6 & \bf 75.8 & 77.4\\ 	
	& \bf BNF-SB-SM & 82.2 & \bf 73.1 & \bf 85.1 & \bf 73.8 & 76.7 & \bf 90.6 & 83.4 & \bf 89.5 & 51.3 & \bf 86.7 & 61.4 & \bf 86.8 & 83.3 & 81.7 & \bf 82.3 & \bf 67.7 & 81.9 & 58.4 & \bf 82.4 & 75.4 & \bf 77.6\\  \Xhline{4\arrayrulewidth}
	 \multirow{9}{*}{PI-IOU}
	 %& DL-CRF& 46.1 & \bf 28.0 & 48.5 & 54.5 & 45.5 & 57.6 & 34.1 & \bf 47.3 & 19.5 & \bf 61.4 & \bf 41.6 & 42.5 & 34.4 & 61.8 & \bf 62.1 & \bf 22.1 & 50.5 & 41.0 & 61.2 & 31.9 & 44.6\\	
	%& \bf DL-CRF+\HfL (VOC) & \bf 47.5 & 27.6 & \bf 50.4 & \bf 63.5 & \bf 47.7 & \bf 57.9 & \bf 38.7 & 47.2 & \bf 21.1 & 57.3 & 41.2 & \bf 43.7 & \bf 36.0 & \bf 66.4 & 61.1 & 21.3 & \bf 53.9 & \bf 42.1 & \bf 70.9 & \bf 34.6 & \bf 46.5\\ \cline{2-23}
	&  FCN-Softmax & 56.9 & 35.1 & 47.8 & 41.1 & 27.4 & 51.1 & 43.4 & 52.7 & 22.2 & 43.1 & 29.2 & 54.2 & 40.5 & 45.6 & 59.1 & 24.2 & 43.6 & 24.8 & 55.9 & 37.2 & 41.8\\
	& Belief Propagation~\cite{Tappen:2003:CGC:946247.946707}  & 68.0 & 38.6 & 52.9 & 45.8 & 31.9 & 55.9 & 47.2 & 58.2 & 24.6 & 49.9 & 31.7 & 60.2 & 44.9 & 50.1 & 62.4 & 25.2 & 49.9 & 27.6 & 62.3 & 42.2 & 46.2\\
	&  ICM  & 65.3 & 40.9 & 56.4 & 45.3 & 33.7 & 58.9 & 49.5 & 61.9 & 25.8 & 53.5 & 33.2 & 62.1 & 48.0 & 53.2 & 63.4 & 24.1 & 54.8 & 34.0 & 63.7 & 47.7 & 48.8\\
	& TRWS~\cite{TRWS} & 67.5 & 40.7 & 60.3 & 46.3 & 35.6 & 63.4 & 49.6 & 69.3 & 29.7 & 58.9 & 37.8 & 67.4 & 57.3 & 53.8 & 64.1 & 26.3 & 62.0 & 36.9 & 63.1 & 49.9 & 52.0\\
	& Graph Cuts~\cite{Boykov:2001:FAE:505471.505473} & 72.1 & 47.8 & 64.5 & 50.8 & 36.0 & 70.8 & 51.4 & 71.6 & 31.7 & 65.8 & 34.4 & 71.8 & 62.0 & 59.4 & 64.8 & 29.0 & 60.9 & 38.7 & 70.3 & 51.6 & 55.3\\
	& QPBO~\cite{QPBO} & 71.6 & 46.8 & 65.6 & 49.6 & 38.0 & 72.6 & 52.7 & 76.7 & 32.5 & 69.6 & 38.9 & 74.4 & 61.4 & 61.0 & \bf 66.2 & 30.3 & \bf 68.7 & \bf 41.4 & 72.2 & 52.8 & 57.2\\
	&  Dense-CRF~\cite{NIPS2011_4296}  & 68.0 & 39.5 & 58.0 & 45.0 & 33.4 & 62.8 & 47.7 & 66.0 & 29.4 & 60.9 & 36.0 & 68.5 & 54.6 & 51.4 & 63.7 & 28.3 & 57.6 & 37.1 & 65.9 & 48.2 & 51.1\\
	&  \bf BNF-SB & 71.6 & 48.1 & \bf 67.2 &  52.3 &  37.8 &  \bf 79.5 &  52.9 &  \bf 80.8 &  \bf 33.3 &  71.5 &  \bf 39.5 &  \bf 75.1 &  65.7 &  63.4 & 65.1 &  31.1 &  67.5 & 39.6 &  \bf 73.2 &   \bf 54.7 &  \bf 58.5\\ 
	&  \bf BNF-SB-SM & \bf 72.0 & \bf 48.9 & 66.5 &  \bf 52.9 &  \bf 39.1 &  79.0 &  \bf 53.4 &  78.6 &  32.9 &  \bf 72.2 &  39.4 &  74.6 &  \bf 65.9 &  \bf 64.2 & 65.8 &  \bf 31.7 &  66.9 & 39.0 &  73.1 &   53.9 &  \bf 58.5\\ \hline

      \end{tabular}
    \end{center}
    \caption{Semantic segmentation results on the SBD dataset according to PP-IOU (per pixel) and PI-IOU (per image) evaluation metrics. We use BNF-SB to denote the variant of our method that uses only semantic boundary based affinities. Additionally, we use BNF-SB-SM to indicate our method that uses boundary and \textit{softmax} based affinities (See Eq.~\eqref{eq:aff}). We observe that our proposed globalization method outperforms other globalization techniques according to both metrics by at least $0.3\%$ and $1.3\%$ respectively. Note that in this experiment, all of the inference methods use~\textbf{the same FCN unary potentials}. Additionally, for each method except Dense-CRF (it is challenging to incorporate boundary based affinities into the Dense-CRF framework) we use our boundary based affinities, since those lead to better results.}
    \label{pp_iou}
   \end{table*}

\subsection{Comparing Affinity Functions for Semantic Segmentation}
\label{aff_results}

% note that we compare against different globalizations (because we dont change unary potentials)
% mention that for all globalizations we use thes same affinity and the same unary potentials
% mention that 0.3 % corresponds to millions of more correctly labeled pixels

 %\LT{used in~Dense-CRF~\cite{NIPS2011_4296}}

In Table~\ref{affinity_table}, we consider two global models. Both models use the same unary potentials obtained from the FCN \textit{softmax} layer. However, the first model uses the popular color-based pairwise affinities, while the second employs our boundary-based affinities. Each of these two models is optimized using several inference strategies. The table shows that using our boundary based-affinity function improves the results of all global inference methods according to both evaluation metrics. Note that we cannot include Dense-CRF~\cite{NIPS2011_4296} in this comparison because it employs an efficient message-passing technique and integrating our affinities into this technique is a non-trivial task. However, we compare our method with Dense-CRF in Subsection~\ref{inf_results}. 

The results in Table~\ref{affinity_table} suggest that our semantic boundary based pixel affinity function yields better semantic segmentation results compared to the commonly-used color based affinities. We note that we also compared the results of our inference technique using other edge detectors, notably UCM~\cite{Arbelaez:2011:CDH:1963053.1963088} and \HfL~\cite{gberta_2015_ICCV}. In comparison to UCM edges, we observed that our boundaries provide $1.0\%$ and $6.0\%$ according to both evaluation metrics respectively. When comparing our boundaries with \HfL method, we observed similar segmentation performance, which suggests that our method works best with the high quality semantic boundaries.

\subsection{Comparing Inference Methods for Semantic Segmentation}
\label{inf_results}

%We present two tables to validate our claims. 

Additionally, we also present semantic segmentation results for both of the metrics (PP-IOU and PI-IOU) in Table~\ref{pp_iou}. In this comparison, all the techniques use the same FCN unary potentials. Additionally, all inference methods except Dense-CRF use our affinity measure (since the previous analysis suggested that our affinities yield better performance). We use BNF-SB to denote the variant of our method that uses only semantic boundary based affinities. Additionally, we use BNF-SB-SM to indicate the version of our method that uses both boundary and \textit{softmax}-based affinities (see Eq.~\eqref{eq:aff}).  

Based on these results, we observe that our proposed technique outperforms all the other globalization methods according to both metrics, by $0.3\%$ and $1.3\%$ respectively. Additionally, these results indicate that most benefit comes from the semantic boundary affinity term rather than the \textit{softmax} affinity term.

In Fig.~\ref{SS}, we also present qualitative semantic segmentation results. Note that, compared to the segmentation output from the \textit{softmax} layer, our segmentation is much better localized around the object boundaries. Additionally, in comparison to Dense-CRF predictions, our method produces segmentations that are much spatially smoother.

%We also note that FCNs make lots of object recognition mistakes~\cite{DBLP:journals/corr/DaiH015}. However, in this work, we focus on. As a result, we note that a more advanced architecture that recognizes objects with higher accuracy could be easily integrated into our architecture and used in the same way as we use our current FCN.

%mention that we can  use any state-of-the-art network for our unary potential rpedictor

\subsection{Semantic Boundary Classification}

We can also label our boundaries with a specific object class, using the same classification strategy as in the \HfL system~\cite{gberta_2015_ICCV}. Since the SBD dataset provides annotations for semantic boundary classification, we can test our results against the state-of-the-art \HfL~\cite{gberta_2015_ICCV} method for this task. Due to the space limitation, we do not include full results for each category. However, we observe that our produced results achieve mean Max F-Score of $54.5\%$ (averaged across all $20$ classes) whereas \HfL method obtains $51.7\%$. 

%We also note that default settings in the benchmark do not consider internal object boundaries for accuracy evaluation, which makes the task slightly easier. In our evaluation, we consider both external and internal object boundaries and show that our method outperforms \HfL.

\section{Conclusions}

In this work we introduced a Boundary Neural Field (BNF), an architecture that employs a semantic segmentation FCN to predict semantic boundaries and then uses the predicted boundaries and the FCN output to produce an improved semantic segmentation maps a global optimization. We showed that our predicted boundaries are better suited for semantic segmentation than the commonly used low-level color based affinities. Additionally, we introduced a global energy function that decomposes semantic segmentation into multiple binary problems and relaxes an integrality constraint. We demonstrated that the minimization of this global energy allows us to predict segmentations that are better localized around the object boundaries and that are spatially smoother compared to the segmentations achieved by prior methods. We made the code of our globalization technique available at {\small \url{http://www.seas.upenn.edu/~gberta/publications.html}}.

The main goal of this work was to show the effectiveness of boundary-based affinities for semantic segmentation. However, due to differentiability of our global energy, it may be possible to add more parameters inside the BNFs and learn them in a front-to-end fashion. We believe that optimizing the entire architecture jointly could capture the inherent relationship between semantic segmentation and boundary detection even better and further improve the performance of BNFs. We will investigate this possibility in our future work.

\section{Acknowledgements}

This research was funded in part by NSF award CNS-1205521.

\bibliographystyle{plain}
\footnotesize{
\bibliography{gb_bibliography}}

\end{document}